\documentclass[letterpaper,10pt,journal,twoside]{IEEEtran}
\usepackage{amsmath,amssymb,stmaryrd,mathtools}
\usepackage[hidelinks]{hyperref}
\usepackage[noabbrev]{cleveref}
\usepackage{cite}
\usepackage{bm}
\usepackage{booktabs}
\usepackage{multirow}
\usepackage{makecell}
\usepackage{xcolor}
\usepackage{paralist}
\usepackage{subcaption}
\usepackage{caption}
\begin{document}

\title{Proactive Anomaly Detection for Robot Navigation with Multi-Sensor Fusion}

\author{Tianchen Ji, Arun Narenthiran Sivakumar, Girish Chowdhary, and Katherine Driggs-Campbell% <-this % stops a space
\thanks{Manuscript received: September 9, 2021; Revised December 10, 2021; Accepted January 23, 2022.}
\thanks{This paper was recommended for publication by Editor Pauline Pounds upon evaluation of the Associate Editor and Reviewers' comments.}
\thanks{T. Ji, A. N. Sivakumar, G. Chowdhary, and K. Driggs-Campbell are with the Coordinated Science Laboratory, University of Illinois at Urbana-Champaign.
        {\tt\footnotesize \{tj12,av7,girishc,krdc\}@illinois.edu}}
\thanks{Digital Object Identifier (DOI): see top of this page.}
}

% The paper headers
\markboth{IEEE Robotics and Automation Letters. Preprint Version. Accepted January, 2022}%
{Ji \MakeLowercase{\textit{et al.}}: Proactive Anomaly Detection for Robot Navigation with Multi-Sensor Fusion}

\maketitle

%%%%%%%%%%%%%%%%%%%%%%%%%%%%%%%%%%%%%%%%%%%%%%%%%%%%%%%%%%%%%%%%%%%%%%%%%%%%%%%%
\begin{abstract}
Despite the rapid advancement of navigation algorithms, mobile robots often produce anomalous behaviors that can lead to navigation failures. The ability to detect such anomalous behaviors is a key component in modern robots to achieve high-levels of autonomy. Reactive anomaly detection methods identify anomalous task executions based on the current robot state and thus lack the ability to alert the robot before an actual failure occurs. Such an alert delay is undesirable due to the potential damage to both the robot and the surrounding objects. We propose a proactive anomaly detection network (PAAD) for robot navigation in unstructured and uncertain environments. PAAD predicts the probability of future failure based on the planned motions from the predictive controller and the current observation from the perception module. Multi-sensor signals are fused effectively to provide robust anomaly detection in the presence of sensor occlusion as seen in field environments. Our experiments on field robot data demonstrates superior failure identification performance than previous methods, and that our model can capture anomalous behaviors in real-time while maintaining a low false detection rate in cluttered fields. Code, dataset, and video are available at \url{https://github.com/tianchenji/PAAD}.
\end{abstract}

%%%%%%%%%%%%%%%%%%%%%%%%%%%%%%%%%%%%%%%%%%%%%%%%%%%%%%%%%%%%%%%%%%%%%%%%%%%%%%%%


%%%%%%%%%%%%%%%%%%%%%%%%%%%%%%%%%%%%%%%%%%%%%%%%%%%%%%%%%%%%%%%%%%%%%%%%%%%%%%%%
\section{Introduction}
\begin{figure}[t]
  \centering
  \begin{subfigure}[b]{0.402\linewidth}
    \includegraphics[width=\linewidth]{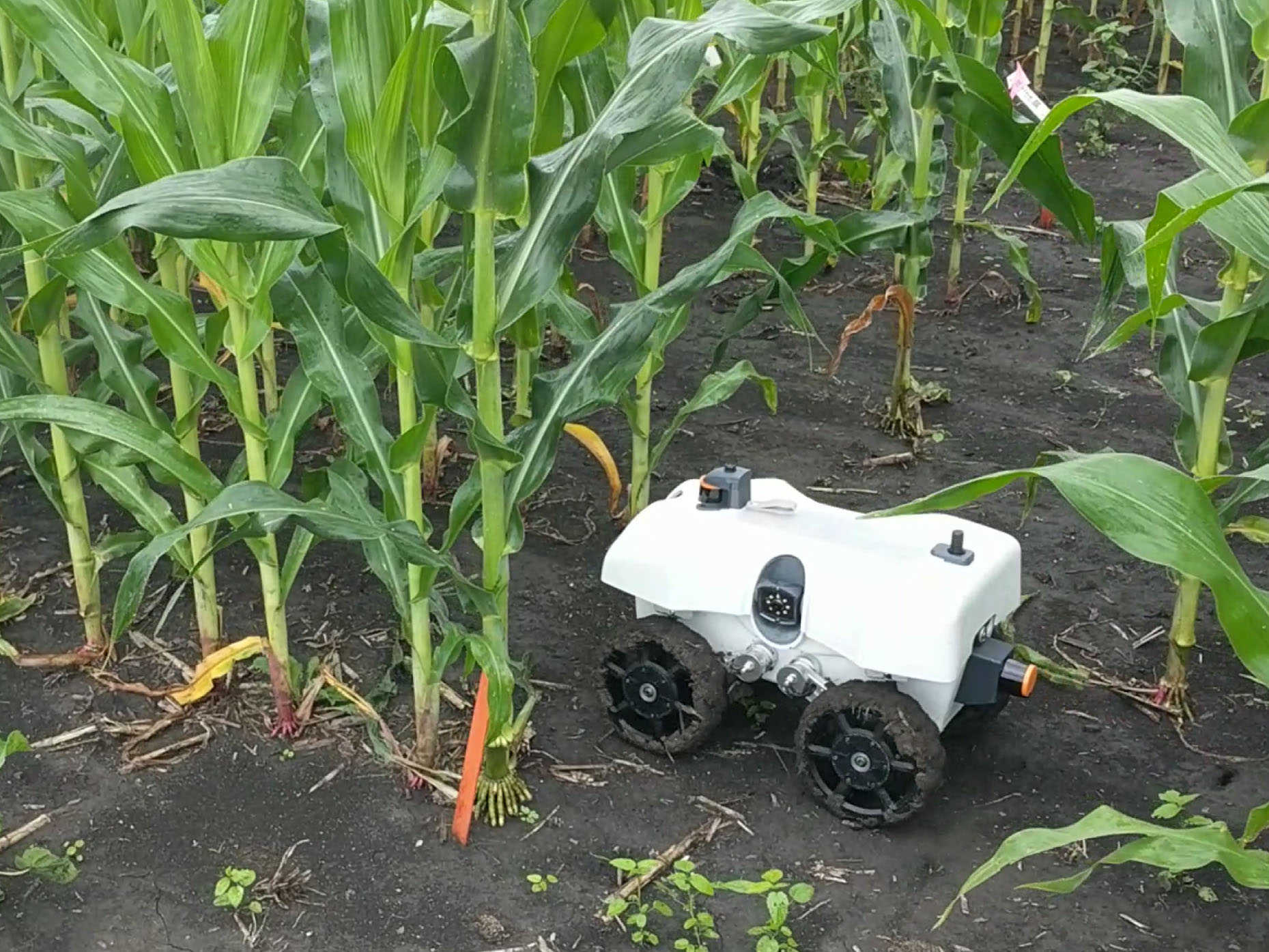}
    \caption{}
    \label{subfig:terrasentia}
  \end{subfigure}
  \begin{subfigure}[b]{0.402\linewidth}
    \includegraphics[width=\linewidth]{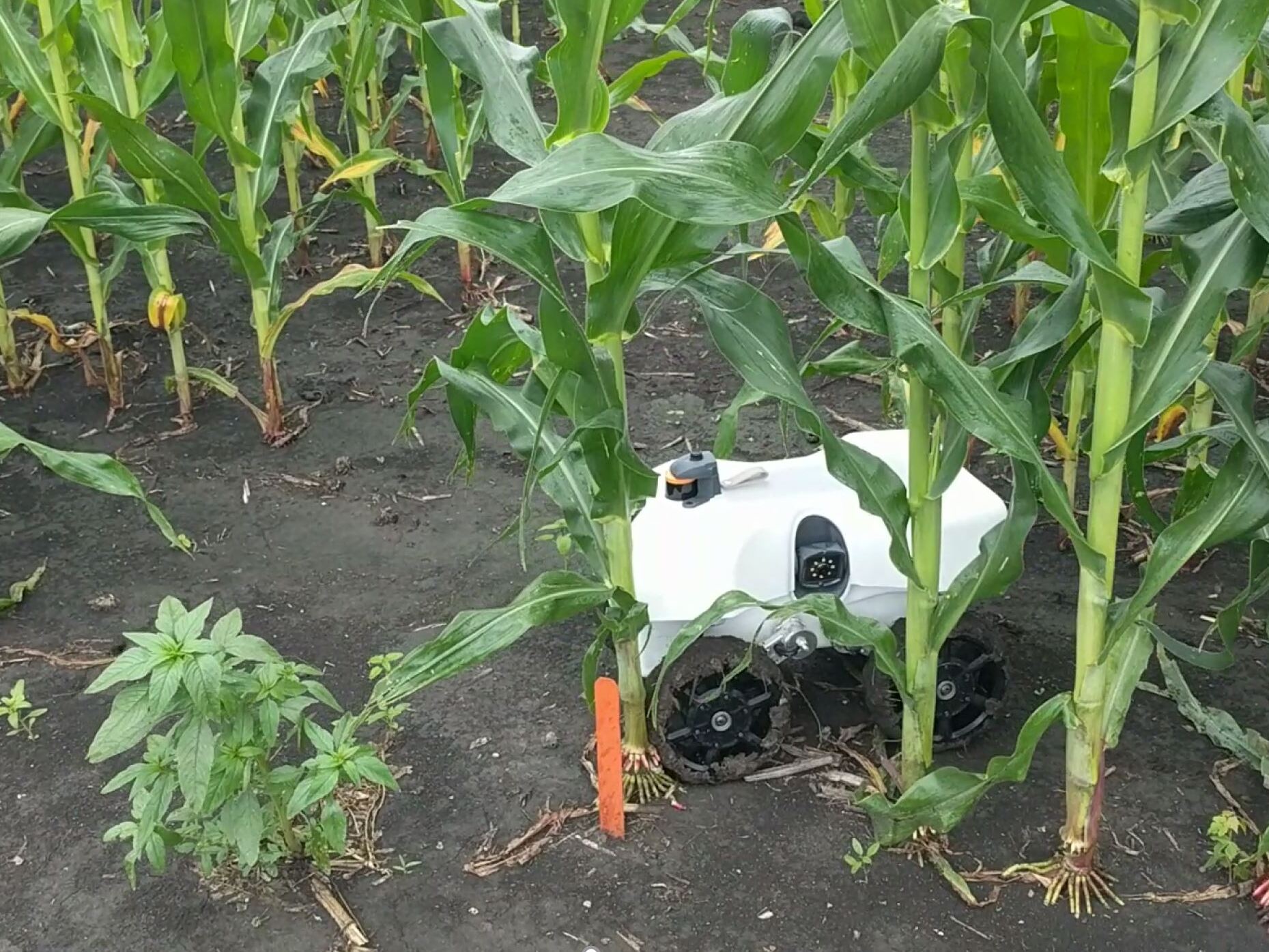}
    \caption{}
    \label{subfig:common-failure}
  \end{subfigure}
  \begin{subfigure}[b]{0.17\linewidth}
    \includegraphics[width=\linewidth]{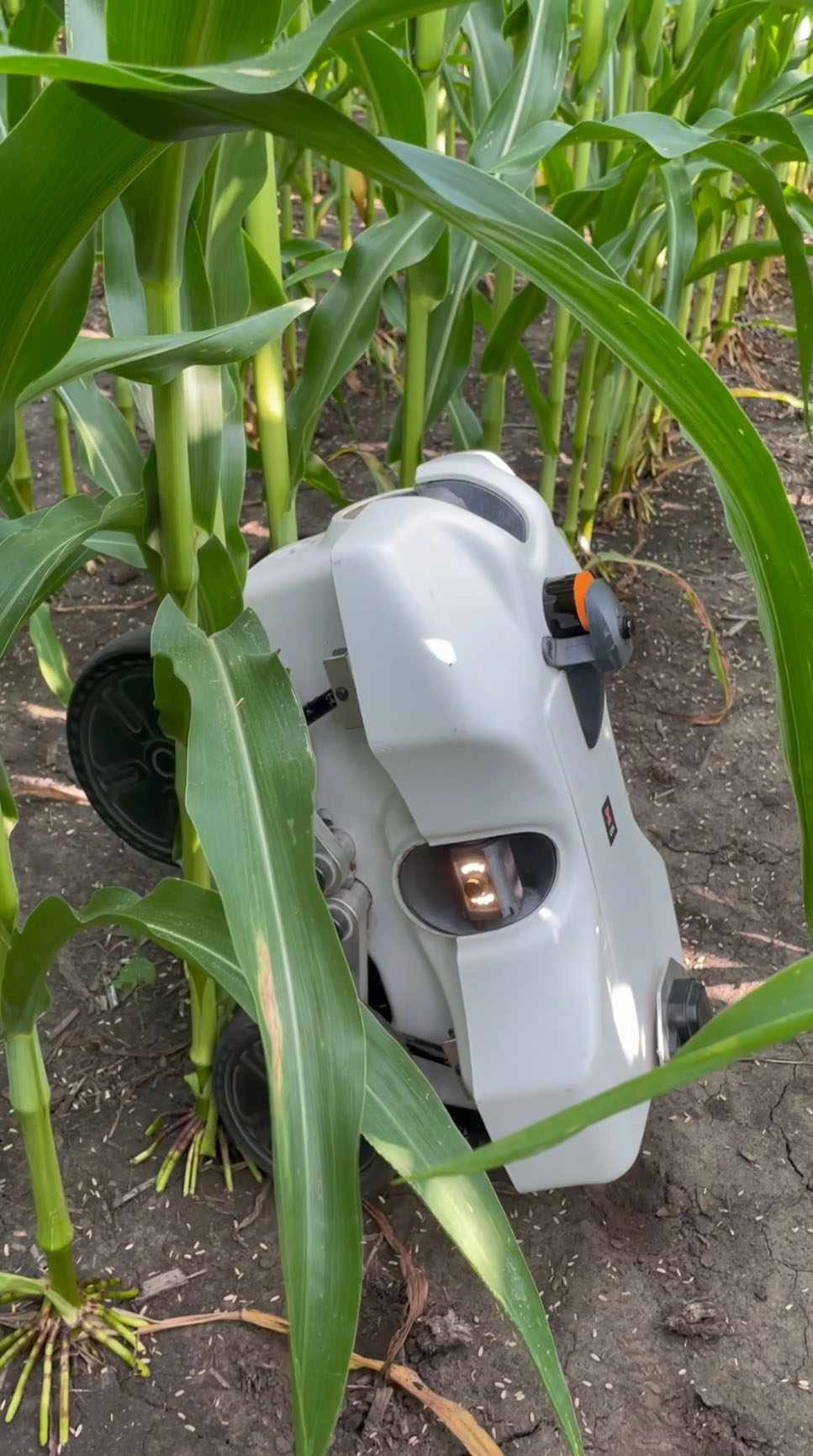}
    \caption{}
    \label{subfig:catastrophic-failure}
  \end{subfigure}
  \caption{\textbf{Field robot platform.} \textbf{(a)} The field robot, TerraSentia, navigates between rows of crops under cluttered canopies to collect data on plant traits. \textbf{(b)} The robot fails the navigation task due to anomalous behaviors. \textbf{(c)} Rare but catastrophic failures may occur in extreme cases.}
  \label{fig:robot-platform}
  \vspace{-3mm}
\end{figure}

\IEEEPARstart{M}{obile} robots are playing an important role in creating intelligent, productive, and easy-to-operate modern farms. Small and low-cost robots (Figure~\ref{subfig:terrasentia}) deployed under crop canopies can increase agricultural sustainability by performing tasks that cannot be accomplished by over-canopy large equipment~\cite{xu2018development}.
%, such as high-throughput plant phenotyping, ultra-precise pesticide treatments, and mechanical weeding~\cite{mueller2017robotanist,xu2018development,mcallister2020agbots,foley2011solutions}.
Although recent research efforts have made noteworthy progress on developing trustworthy autonomy for robot navigation~\cite{kayacan2018embedded,zhang2020high,velasquez2021multi,sivakumar2021learned}, robots may still fail out in the field due to the environmental complexity, terrain variability, and sensor uncertainty in field environments. A lack of a detection system for anomalous behaviors before failures may cause damage to robots and plants due to collisions. The detection of such anomalous behaviors can stop the robot from entering failure modes, thus providing opportunities for executing recovery maneuvers and proceeding with the task.

Deep-learning based anomaly detection (AD) algorithms have been widely adopted in robotic applications~\cite{chalapathy2019deep}. Many previous works approached the AD problem in a reactive manner~\cite{ji2020multi,park2018multimodal,park2019multimodal,malhotra2016lstm} (i.e., an anomaly is detected when the current system state reveals a different pattern from that of past successful experiences). Such reactive anomaly detectors make an inference merely based on the current sensory signals (e.g., velocity, torque, LiDAR readings) and lack the ability to predict potential failures in the future. As a result, the robot may still be damaged due to collisions (Figure~\ref{subfig:common-failure}) or enter critical states (Figure~\ref{subfig:catastrophic-failure}), the recovery from which is beyond the robot autonomy, due to the alert delay.

\begin{figure}[t]
  \centering
  \begin{subfigure}[b]{0.48\linewidth}
    \includegraphics[width=\linewidth]{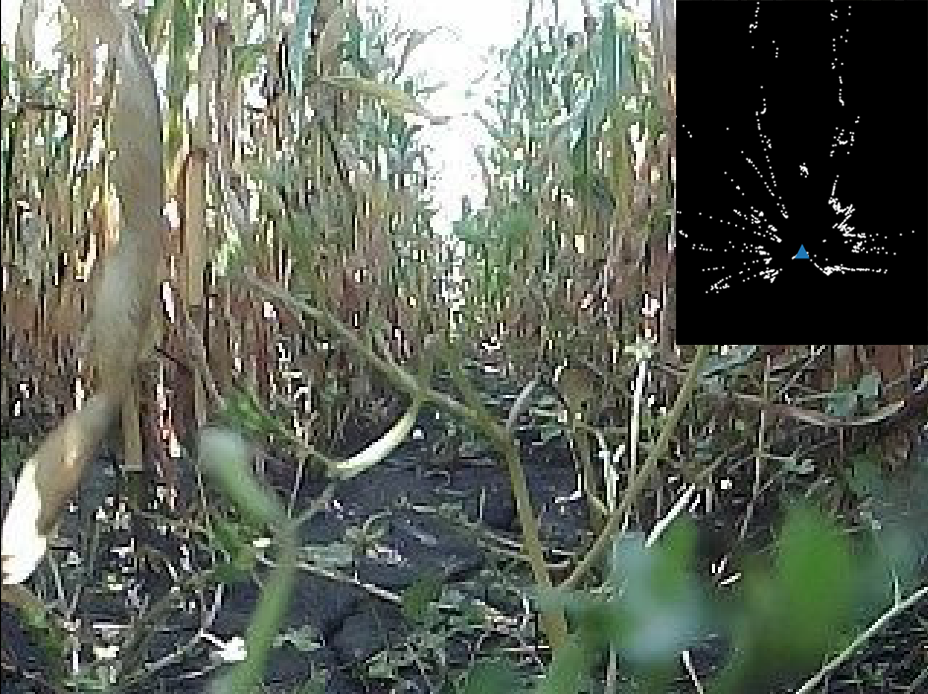}
    \caption{}
    \label{subfig:noisy-environment}
  \end{subfigure}
  \begin{subfigure}[b]{0.48\linewidth}
    \includegraphics[width=\linewidth]{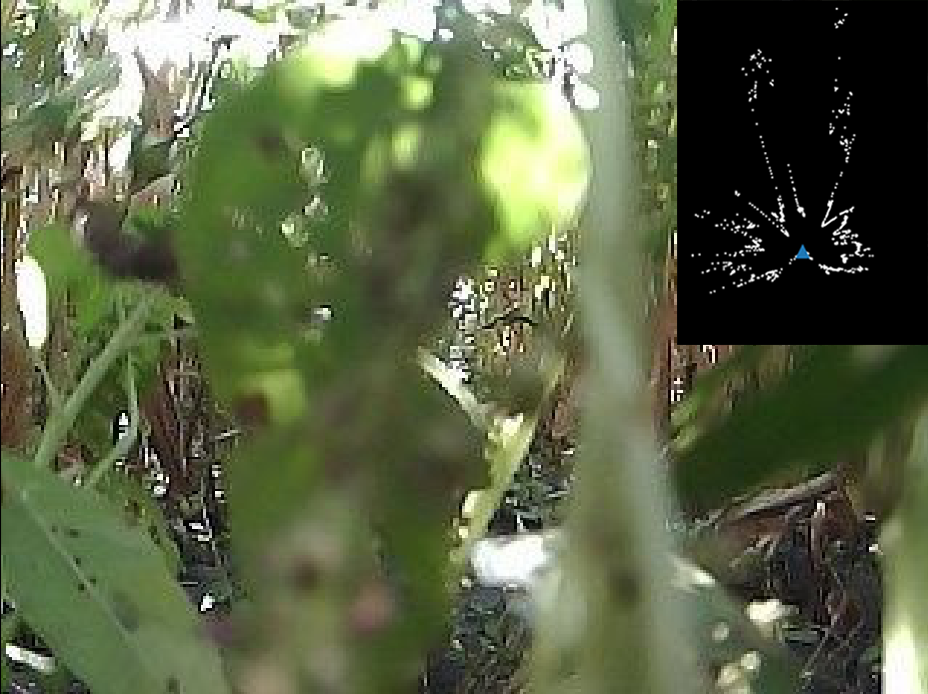}
    \caption{}
    \label{subfig:occlusion}
  \end{subfigure}
  \caption{\textbf{Field environment.} The robot perceives the environment through a forward-facing camera and a 2D LiDAR. The blue triangle in the 2D point cloud denotes the robot. Weeds and low-hanging leaves introduce high uncertainty in sensory signals and can block the sensor view as the robot navigates under canopy.}
  \label{fig:field-environment}
  \vspace{-3mm}
\end{figure}

An alternative solution is \textit{proactive} anomaly detection, which predicts the probability of future failure based on the planned actions and the current sensory observation. Such predictive model has been explored in LaND~\cite{kahn2021land} and BADGR~\cite{kahn2021badgr} to choose optimal actions for outdoor navigation.  However, the AD problem for robot navigation through natural field environments introduces challenges which are usually not considered when deploying autonomous systems in common outdoor environments. First, the perception system and control system both exhibit high uncertainty during operation. Useful features for AD tasks (e.g., relative position of the robot with respect to the crop rows) are usually buried in noisy sensory signals due to weeds, lodged plants, and low-hanging leaves (Figure~\ref{subfig:noisy-environment}). Meanwhile, the actions (e.g., linear and angular velocity) executed by the robot are constantly corrupted by varying wheel-terrain interactions~\cite{kayacan2018embedded}, introducing high variance in control signals that can be problematic for pattern recognition. Second, the frequent sensor occlusion imposes challenges on the robot perceiving the environment (Figure~\ref{subfig:occlusion}). Anomaly detectors relying on single sensor modality~\cite{kahn2021land,kahn2021badgr,ji2020multi} can be easily fooled due to the lack of a robust perception system.

In this paper, we approach the proactive anomaly detection problem by identifying anomalous behaviors conditioned on the current observations. Formally, we define an anomalous behavior for robot navigation as a sequence of future motions which contains at least one time step with failure within the prediction horizon. Such future motion can be represented as a set of control actions or a planned path.

We introduce a \textbf{P}ro\textbf{A}ctive \textbf{A}nomaly \textbf{D}etection network, which we call PAAD, that reasons about probability of failure at each time step within the future time horizon by leveraging the planned motions from the predictive controller and the current observation from the perception system.
%We employ a low-variance image representation of planned motions and take advantage of multi-sensor signals for robust perception.
Features from different modalities are extracted independently and fused in two stages to generate the final probability of failure. We train PAAD with a mixed cost function, consisting of a prediction task and a reconstruction task, to improve the generalization capability and increase the robustness against noisy sensory signals.

Our contributions can be summarized as follows:
\begin{inparaenum}[(1)]
\item
We propose a novel deep neural network architecture called PAAD, which effectively fuses multi-sensor signals for robust perception in unstructured and uncertain environments.
\item
We employ a low-variance image representation of planned motions, as opposed to raw control actions, to realize \textit{proactive} anomaly detection and to facilitate efficient feature extraction from noisy signals.
\item
Our proposed detector outperforms existing methods in failure identification performance on an offline real-world navigation dataset and is able to catch anomalous behaviors online while maintaing a low false detection rate in a real-time test.
\end{inparaenum}

%We evaluated PAAD on $4.1$ km of real-world navigation data collected by a TerraSentia robot in corn fields. We demonstrate that the proposed detector outperforms existing methods in failure identification performance. In our real-time test on an additional $1.3$ km of field navigation, we further show that PAAD is able to catch anomalous behaviors online while maintaining a low false detection rate.
\section{Related Work}
Anomaly detection, also known as outlier detection or novelty detection, is an important problem that has been studied within diverse research areas and application domains~\cite{chandola2009anomaly,chalapathy2019deep}. In robotics, AD has been used to detect failures of manipulation tasks~\cite{park2016multimodal,kappler2015data} and navigation tasks~\cite{wellhausen2020safe,ji2020multi}.

% Early research efforts often make use of system models and/or sensor models for robot AD tasks~\cite{schneider1994fuzzy,goel2000fault,vemuri1997neural}. Schneider et al. compares the actual states to the reference or simulated states and declares an anomaly if the residual exceeds a predefined threshold~\cite{schneider1994fuzzy}. Multiple Model Adaptive Estimation (MMAE) extends the idea to multi-class anomaly detection by using a bank of Kalman filters to predict the outcome of several failure patterns. A neural network is then trained to identify which failure has occurred based on the residual between the predicted readings and the actual sensor readings~\cite{goel2000fault}.  These approaches have been shown effective in detecting anomalies which can be modeled in system dynamics; however, the generalization of these model-based methods to complex environment with high uncertainty remains unclear.

%Recently, learning-based AD algorithms have become increasingly popular.
Recent research efforts have made noteworthy progress in developing learning-based AD algorithms. Maalhotra et al. introduces an LSTM-based encoder-decoder scheme for multi-sensor anomaly detection (EncDec-AD) that uses reconstruction error to detect anomalies~\cite{malhotra2016lstm}. Park et al. proposes a multimodal LSTM-based variational autoencoder (LSTM-VAE) that fuses sensory signals and reconstructs their expected distribution. A reconstruction-based anomaly score
%and a state-based threshold are
is then used to detect anomalies~\cite{park2018multimodal}. Our previous work casts the AD problem as a multi-class classification problem and proposes the use of a supervised variational autoencoder model (SVAE)~\cite{ji2020multi}.
%, which utilizes the representational power of VAE to extract robust features from high-dimensional inputs for the classification task.
However, these reactive methods lack the ability to detect anomalous behaviors before failures and thus the safety is not necessarily enhanced~\cite{hornung2014model}.

In the domain of proactive anomaly detection / predictive collision avoidance, the most similar work to PAAD is the predictive model for future navigational events (e.g., collision) proposed in LaND~\cite{kahn2021land} and BADGR~\cite{kahn2021badgr}. The neural network takes as input an image and a sequence of future control actions, and predicts the probabilities of collision for each time step within the prediction horizon. The model has been shown to have reliable anomaly detection performance on sidewalks and off-road environments with large open space. However, such network suffers from sensor occlusion due to the unimodal input and can struggle with learning useful features due to the input uncertainty. In this work, we make use of both camera and LiDAR data to improve the robot perception capability, and use the image representation of the planned path rather than noisy control actions to facilitate the efficient training of the model.

Another widely explored research area that is relevant to our work is traversability analysis in unstructured environments. Terrain traversability analysis can be referred to as the problem of estimating the difficulty of driving through a terrain for a ground vehicle~\cite{guastella2021learning}. Bekhti et al. use terrain images and acceleration signals to train a Gaussian process regressor in order to predict vibrations using only image texture features~\cite{bekhti2020regressed}. Maturana et al. propose a real-time mapping strategy that provides a 2.5D grid map centered on the vehicle frame, encoding both geometry and semantic information of the environment~\cite{maturana2018real}. Despite their similarity in methodology, traversability estimation and anomaly detection aim at fundamentally different tasks: navigating over a traversable terrain does not imply that the robot behavior is not anomalous. In field environments, for example, a trajectory that drives off the trail from one to another due to large gaps between crops can be collision free and incur no additional traversal cost; however, such behavior should be classified as an anomaly as the robot is deviating from the specified navigation task.

As an emerging research theme, the camera-lidar fusion has been applied to diverse research areas in robotics and autonomous driving~\cite{cui2021deep}. Typical applications include depth completion~\cite{cheng2020cspn++}, object detection~\cite{liang2018deep}, object tracking~\cite{zhang2019robust}, and simultaneous localization and mapping (SLAM)~\cite{xu2018slam}. However, one typical assumption that these application domains make is that the camera and LiDAR data are consistent (i.e., the perceived worlds from the two modalities can be matched with each other). The cluttered environment in fields breaks such assumption as one of the sensors can be occluded frequently and thus poses extra challenges in applying previous techniques. In agricultural settings, the perception error from sensor occlusion is often viewed as noise and is handled by Kalman filter~\cite{velasquez2021multi,sivakumar2021learned}. Although such filtering approach can overcome the problem of occlusion to some extent, the required assumption (e.g., the center line is free of obstacles) does not always hold in the real world. In this work, we develop a novel sensor fusion mechanism to combat sensor occlusion in cluttered environments.

\begin{figure*}[t]
  \centering
  \includegraphics[width=0.98\linewidth]{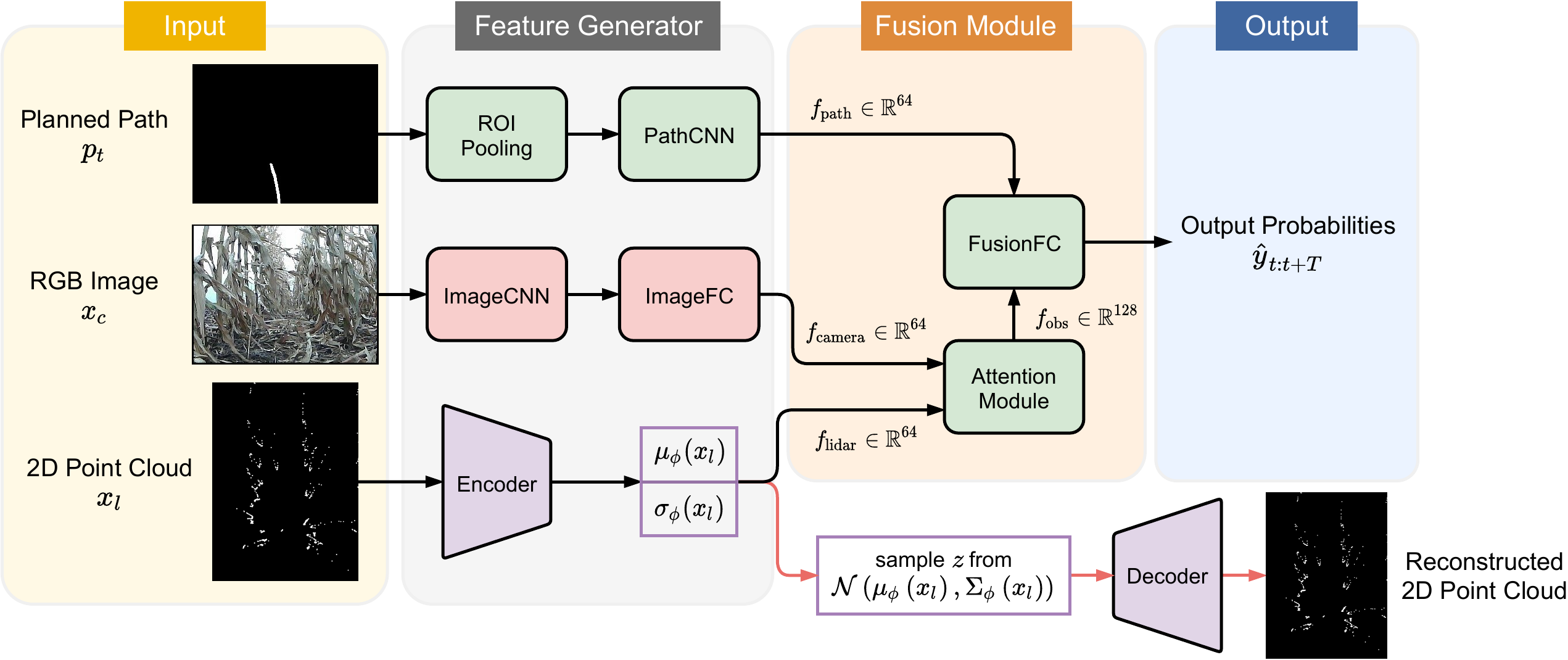}
  \caption{\textbf{Model architecture overview.} The planned path, camera image and LiDAR point cloud are fed to parallel feature generators, then fused in two stages to learn the probability of failure within the prediction horizon. The processing pipelines for the planned path, camera image, and LiDAR point cloud are in green, red, and purple, respectively. PAAD in test time is highlighted in background colors while red arrows are used for training only.}
  \label{fig:PAAD}
  \vspace{-2mm}
\end{figure*}
\section{Method}
Our goal is to develop an AD module that enables a mobile robot to detect anomalous behaviors proactively during navigation in the field. 

 We assume that the sensor observations at time $t$, $\mathbf{o}_t$, are multi-modal, consisting of an RGB image $\mathbf{x}_c \in \mathbb{R}^{H \times W \times C}$ from camera and range readings $\mathbf{x}_l \in \mathbb{R}^L$ from 2D LiDAR. The robot employs a predictive controller which plans a sequence of actions for the next $T$ time steps in a receding horizon manner. We further retrieve the current planned path from subsequent actions and represent the resulting path as a separate image $\mathbf{p}_t \in \mathbb{R}^{H \times W \times 1}$, in which the path is projected onto a blank front-view image plane. At each time step, the task for the AD module is to map from a set of current sensor observations and a planned path $(\mathbf{o}_t, \mathbf{p}_t) \in \mathbb{R}^{H \times W \times C} \times \mathbb{R}^L \times \mathbb{R}^{H \times W \times 1}$ to a sequence of probabilities of navigation failure $\hat{\mathbf{y}}_{t:t+T} \coloneqq (y_t, y_{t+1}, \dots, y_{t+T-1})$ for each time step along the path as shown in Figure~\ref{fig:PAAD}.

Compared to the existing reactive anomaly detection method, PAAD is able to make use of the modality from the planning module to detect anomalous behaviors. Such a proactive nature of PAAD alerts the robot before entering critical states from which human interventions are required to recover the robot. Furthermore, the effective fusion of multi-modal perception signals provides robustness against uncertainty and sensor occlusion in complex field environments. By contrast, false detection of an anomaly can be triggered frequently due to camera occlusion in anomaly detectors that use unimodal perception signals~\cite{kahn2021land,kahn2021badgr}. Lastly, the adopted image representation of the planned path possesses less variance than the raw control actions, thus leading to a more efficient training procedure.

In the following sections, we will describe the data collection process, model architecture, and training procedure.

\subsection{Data Collection}
\label{subsec:data-collection}
The TerraSentia robot is an ultra-compact 4-wheeled skid-steering mobile robot designed to drive through crop rows for automated phenotyping~\cite{kayacan2018embedded}. The robot is equipped with a forward facing monocular camera sensor (OV2710) and a 2D horizontal-scanning LiDAR (Hokuyu UST-10LX) which covers a $270^{\circ}$ range with $0.25^{\circ}$ angular resolution. The observation $\mathbf{o}_t$ is defined by a $240 \times 320$ RGB image and a vector of LiDAR ranges of dimension 1081. The image representation of the planned path $\mathbf{p}_t$ is generated from the output of onboard model predictive controller using perspective projection. The ground truth probability of failure $y_t$ is a binary number indicating if the robot is in a normal state or fails the navigation task.

During Data collection, the robot executes an autonomous control policy: in our case, the LiDAR-based navigation algorithm for agricultural mobile robots~\cite{higuti2019under}. Once the robot enters a failure mode, the human disengages the autonomy, repositions the robot to the center line, and then reactivates the autonomy. We define the failure mode as any state upon entering which the robot is not able to continue the specified navigation task (e.g., following a crop row) without human intervention.

The robot collects the observations, planned paths, and drive modes $(\mathbf{o}_t, \mathbf{p}_t, y_t)$ at each time step $t$. We note that PAAD does not require any additional data beyond what is typically collected for testing the robot autonomy. In fact, the data collection process described above is not dedicated to PAAD but to the testing of LiDAR-based autonomy for agricultural robots~\cite{higuti2019under,velasquez2021multi}.

\subsection{Model Architecture}
We denote PAAD as a function $g: (\mathbf{o}_t, \mathbf{p}_t) \mapsto \hat{\mathbf{y}}_{t:t+T}$, which takes as input a set of current observations and a planned path $(\mathbf{o_t}, \mathbf{p}_t)$ and outputs a sequence of probabilities of failure $\hat{\mathbf{y}}_{t:t+T}$ within the prediction horizon.

The network structure is shown in Figure~\ref{fig:PAAD}. Feature generators (FGs) are designed independently for each modality to extract robust features from different inputs. To strengthen the perception capability in harsh and cluttered agricultural fields, we adopt feature-level camera-lidar fusion, as opposed to signal-level fusion which can struggle with inconsistency in perception signals due to frequent occlusion of one of the sensors. As the final output, the probabilities of navigation failure in the next $T$ time steps are evaluated on the planned path, conditioned on the current observations.

\subsubsection{Feature Generator}
The planned path and RGB image are processed by two separate convolutional pipelines to generate \textit{path features} $f_{\text{path}}$ and \textit{camera features} $f_{\text{camera}}$, respectively. Each CNN module is followed by a flattening operation. For the path image, we crop according to the region of interest (ROI) so that the model is not provided non-essential data which do not include the actual path (e.g., the pixels above the horizon line are always in black).

To extract features from LiDAR point cloud, we borrow the idea from SVAEs~\cite{ji2020multi}. The reconstruction task in the LiDAR pipeline serves as a regularization~\cite{le2018supervised,liu2016algorithm,caruana1997multitask}, which forces the encoder to learn representative features of high-dimensional LiDAR data that are critical to both the downstream inference and generative model. With additional attention on the reconstruction task, the model tends to improve the generalization performance on the inference task~\cite{ji2020multi}. We approximate the posterior distribution of the latent variable $\mathbf{z} \in \mathbb{R}^d$ as a Gaussian with variational parameters $\phi$:
\begin{equation}
q_\phi(\mathbf{z} | \mathbf{x}_l)
=
\mathcal{N} (\mathbf{z} \, | \, \bm{\mu}_\phi (\mathbf{x}_l), \text{diag}(\bm{\sigma}_\phi(\mathbf{x}_l))),
\end{equation}
where $\bm{\mu}_\phi(\mathbf{x}_l)$ is a mean vector, $\bm{\sigma}_\phi(\mathbf{x}_l)$ is a variance vector, and the nonlinear transformations $\bm{\mu}_\phi:\mathbb{R}^L \mapsto \mathbb{R}^d$ and $\bm{\sigma}_\phi:\mathbb{R}^L \mapsto \mathbb{R}^d$ are parameterized by multilayer perceptrons (MLPs) in the encoder.

For the downstream prediction task, we choose \textit{LiDAR features} as:
\begin{equation}
f_\text{lidar} = [\bm{\mu}_\phi (\mathbf{x}_l), \bm{\sigma}_\phi (\mathbf{x}_l)].
\end{equation}
For the reconstruction task~\footnote{During test time, the reconstruction branch is abandoned and \textit{LiDAR features} $f_\text{lidar}$ are forwarded to the fusion module for the prediction task.}, the decoder uses a generative model of the form:
\begin{equation}
p_\theta(\mathbf{x}_l | \mathbf{z})
=
\mathcal{N} (\mathbf{x}_l \, | \, \text{MLP}(\mathbf{z};\theta), \sigma^2 \cdot I),
\end{equation}
where $\text{MLP}(\mathbf{z};\theta)$ is a mean vector formed by a nonlinear transformation of the latent variable $\mathbf{z}$, and $\sigma$ is a hyperparameter. Here, we choose the nonlinear transformation to be an MLP parameterized by $\theta$. Note that the reconstruction branch in LiDAR pipeline follows the structure of a vanilla variational autoencoder (VAE).

\subsubsection{Fusion Module}
To form \textit{observation features} from sensors, we employ a feature-level camera-lidar fusion by using a multi-head attention (MHA) with a residual connection~\cite{he2016deep}:
\begin{equation}
f_\text{obs}
=
[f_\text{camera}, f_\text{lidar}] + \text{MHA}(Q, K, V = [f_\text{camera}, f_\text{lidar}])
\end{equation}
which corresponds to the attention module in Figure~\ref{fig:PAAD}. The query, key, and value are chosen identically to be the concatenation of $f_\text{camera}$ and $f_\text{lidar}$, which can be viewed as a sequence of length $2$. We choose an MHA over an MLP for camera-lidar fusion due to the fact that we expect the model to generate \textit{observation features} based on the signal quality of each sensor. For example, in cases where the camera is blocked by leaves while the LiDAR view is clear, the point cloud should contribute more to \textit{observation features} than the image.

The final fusion of \textit{observation features} and \textit{path features} at time $t$ produces the predicted probability of failure in the next $T$ time steps:
\begin{equation}
\label{eq:fusion-MLP}
\hat{\mathbf{y}}_{t:t + T}
=
\text{Sigmoid}\left(\text{FusionFC}\left([f_\text{obs}, f_\text{path}]\right)\right).
\end{equation}
A sigmoid function is used to ensure that the final output probabilities are scaled into the valid range.

\subsection{Training}
\label{subsec:training}
The ImageCNN in camera pipeline uses a ResNet-18 backbone pretrained on visual navigation task, in which the  network learns to predict robot heading and placement in a crop row using a front-view RGB image~\cite{sivakumar2021learned}. We construct the ImageCNN module by truncating the model of visual navigation right before the fully connected layers. The weights of the ImageCNN are fixed after pretraining.

Denoting the dataset collected in Section~\ref{subsec:data-collection} by $\mathcal{D}$, we specify the overall loss function for PAAD as:
\begin{equation}
\label{eq:loss-function}
\begin{aligned}
\mathcal{L}
=&
\sum_{(\mathbf{o}_t, \mathbf{p}_t, \mathbf{y}_{t:t+T}) \in \mathcal{D}}
\alpha \cdot \mathcal{L}^\text{BCE} (g(\mathbf{o}_t, \mathbf{p}_t), \mathbf{y}_{t:t+T}) \\
&-
\mathbb{E}_{q_\phi (\mathbf{z} | \mathbf{x}_l)} [\log p_\theta (\mathbf{x}_l | \mathbf{z})]
+
D_\text{KL} [q_\phi (\mathbf{z} | \mathbf{x}_l) \| p_\theta (\mathbf{z})],
\end{aligned}
\end{equation}
where $\mathcal{L}^\text{BCE}$ is the binary cross-entropy loss, $\alpha$ is a hyperparameter controlling the relative weight between the discriminative and generative learning, and $p_\theta (\mathbf{z})$ is a prior distribution over the latent variable $\mathbf{z}$. As in SVAEs, we choose $p_\theta (\mathbf{z})$ to be a standard Gaussian distribution $\mathbf{z} \sim \mathcal{N}(0, I)$.

The training objective consists of a prediction task and a reconstruction task. The first term in equation~(\ref{eq:loss-function}) penalizes the prediction error. We set $\alpha=0.1 \cdot N$, where $N$ is the total number of datapoints as in~\cite{ji2020multi}. The last two terms in the loss function, which is also the negative of the evidence lower bound (ELBO) in vanilla VAEs, penalizes the reconstruction error of LiDAR data. The last KL divergence term can be viewed as a regularization.

The inference model and the generative model can be optimized jointly by stochastic gradient descent of the overall objective function~(\ref{eq:loss-function}). To enable the backpropagation through the sampling layer within the network, a common reparameterization trick is used to move the sampling process to a stochastic input layer~\cite{doersch2016tutorial}.
\section{Experimental Results}
In our experiments, we evaluate the anomaly detection performance of PAAD on $4.1$ km of real-world navigation data collected with the TerraSentia robot in corn fields from September to October 2020. The robot navigates between rows of crops under cluttered canopy without damaging the plants. Depending on the environmental conditions, the robot may or may not enter a failure mode in a run. The reference speed for the robot is set to be $0.6$ m/s and the two consecutive points on the planned path from onboard MPC have an interval of $0.2$ meters. After the data collection, we subsample the data to $3$Hz so that the ground truth probability of failure is aligned in time with the predicted one along the planned path.

\begin{figure}[t]
  \centering
  \begin{subfigure}[b]{0.49\linewidth}
    \includegraphics[height=1.225in]{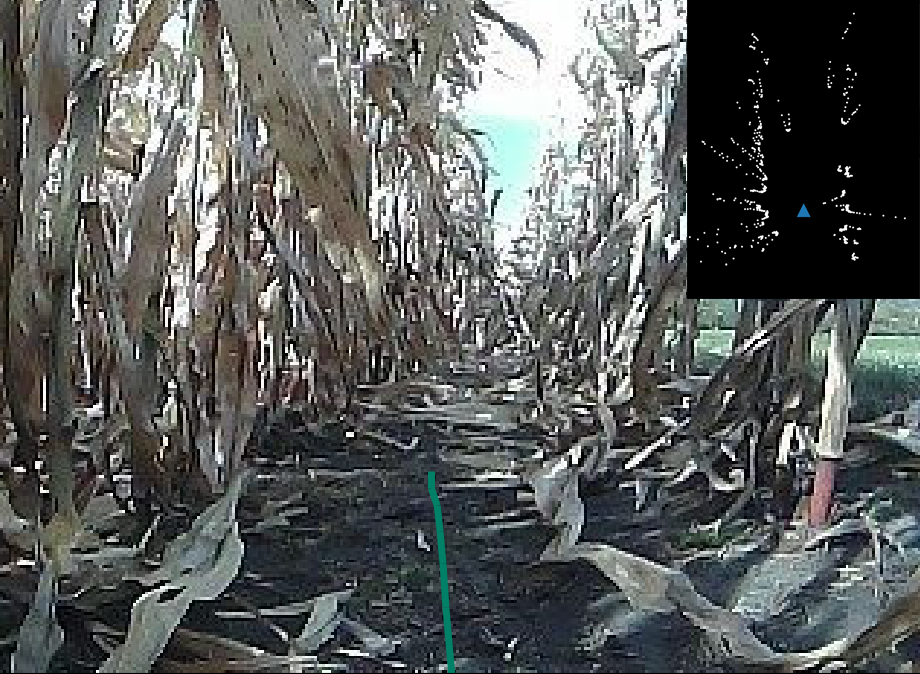}
  \end{subfigure}
  \vspace{1mm}
  \begin{subfigure}[b]{0.49\linewidth}
    \includegraphics[height=1.225in]{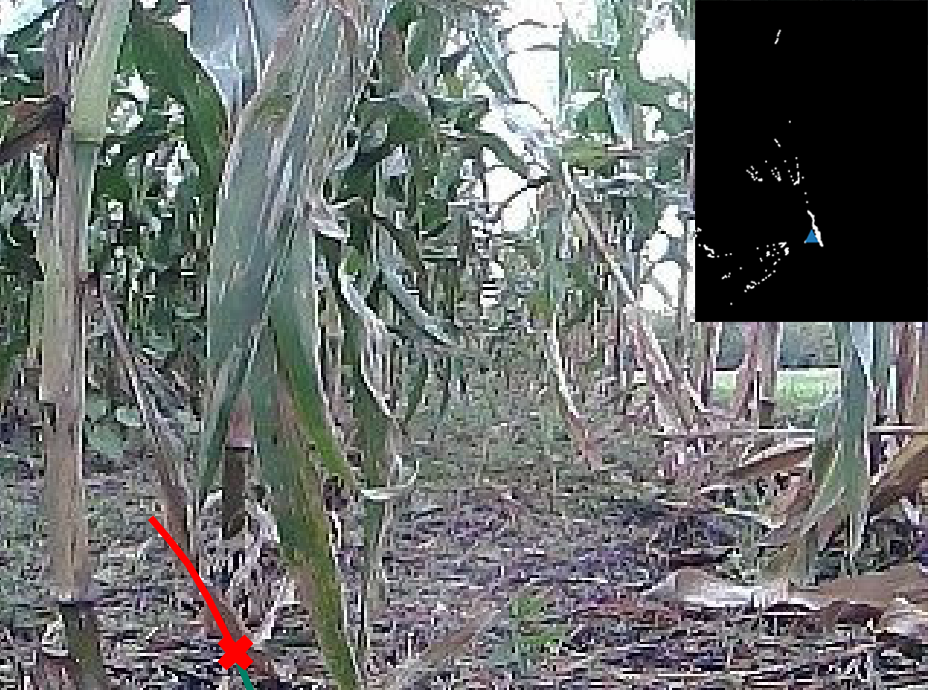}
  \end{subfigure}
  \begin{subfigure}[b]{0.49\linewidth}
    \includegraphics[height=1.24in]{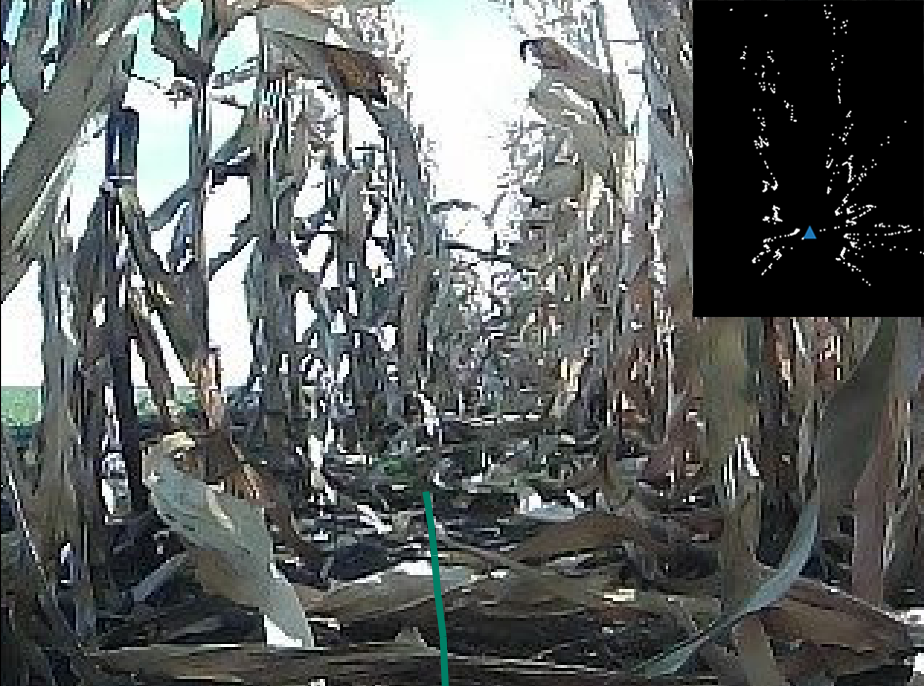}
    \caption{Normal behavior}
  \end{subfigure}
  \begin{subfigure}[b]{0.49\linewidth}
    \includegraphics[height=1.24in]{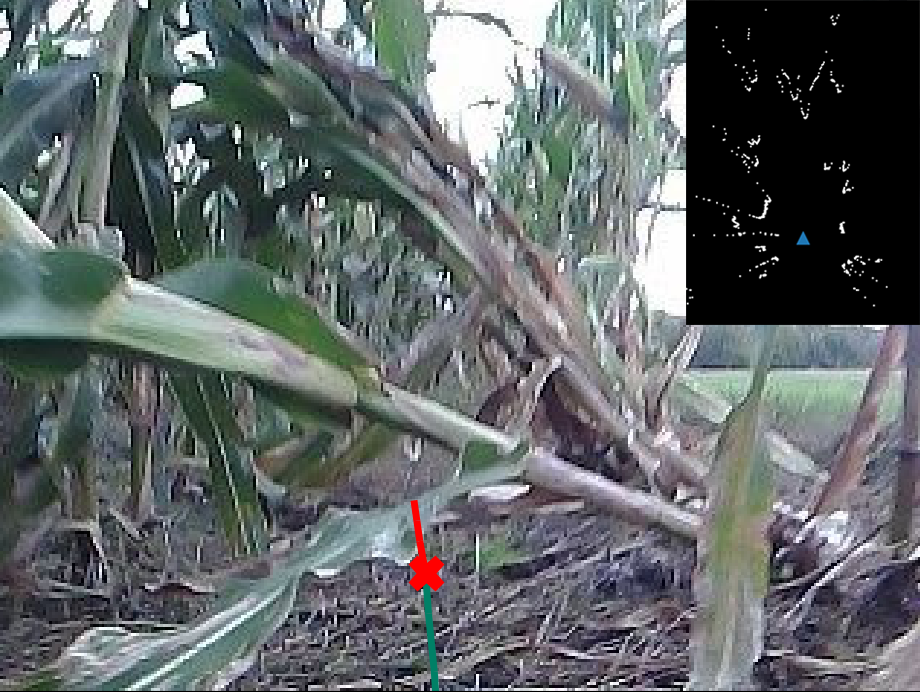}
    \caption{Anomalous behavior}
  \end{subfigure}
  \caption{\textbf{Samples from the collected dataset.} The planned path is overlayed onto the image for visualization purposes. The ground truth probability of failure along the path is indicated by the color (green for normal and red for failure), and the red cross marks the start of a sequence of failures.}
  \label{fig:dataset}
  \vspace{-3mm}
\end{figure}

For all proactive anomaly detectors, we use a prediction horizon of $T=10$ time steps (i.e., a lookahead distance of $1.8$ meters). A subset of our dataset\footnote{The navigation dataset used in this paper can be found at \url{https://github.com/tianchenji/PAAD}} is visualized in Figure~\ref{fig:dataset}. To alleviate the negative effect on the evaluation of different models introduced by the covariance between datapoints closely related in time, we construct the training set and test set from experiments on independent days. The training set consists of $29292$ datapoints and contains $2258$ anomalous behaviors collected over five days, while the test set consists of $6869$ datapoints and contains $689$ anomalous behaviors from data collected on two additional days. The data were collected in part at the Illinois Autonomous Farm. We perform under-sampling of normal cases and over-sampling of anomalous cases on the training set to balance the learning of both types of behaviors while keeping the test set unchanged.

In experiments using PAAD, we construct the PathCNN with 3 convolutional layers with filter number $\{8, 16, 32\}$, filter size $3 \times 3$, and stride $2$. Each convolutional layer is followed by a max pooling layer. We implement the ImageFC with one hidden layer with 64 hidden units. As in SVAEs, the encoder in LiDAR pipeline is constructed by one hidden layer and 128 hidden units, and the decoder mirrors the structure of the encoder. We choose a latent space of dimension $32$ ($\mathbf{z} \in \mathbb{R}^{32}$). In the fusion module, the MHA has $8$ attention heads and the FusionFC has $2$ hidden layers with $\{128, T\}$ hidden units. ReLU activation functions are applied and an Adam optimizer with a constant learning rate of 0.0005 is used to train the network.

\subsection{Baselines and Numerical Evaluation}
\label{subsec:numerical-evaluation}
We evaluate the performance of the proposed method on the test set, along with the following baseline methods:
\begin{itemize}
\item
\textit{CNN-LSTM}: An image-based, action-conditioned convolutional recurrent deep neural network introduced in LaND~\cite{kahn2021land} and BADGR~\cite{kahn2021badgr}. An LSTM unit, initialized by image features generated by a backbone convolutional network, sequentially processes each of the future $T$ control \textit{actions} and outputs the corresponding predicted probability of failure.

\item
\textit{Cui et. al.}~\cite{cui2019multimodal}: A feedforwad convolutional neural network processing an image and robot's \textit{actions} for behavior prediction.

\item
% \textit{PAAD-camera}: We remove the LiDAR pipeline and the attention module of camera-lidar fusion in PAAD. The future probability of failure is only evaluated on an image and a planned \textit{path}.
\textit{NMFNet}~\cite{nguyen2020autonomous}: A multimodal fusion network for robot navigation in complex environments. To evaluate the future probability of failure, we take the two branches that handle LiDAR data and 2D images to process sensor observations and replace the branch of 3D point cloud with an MLP that processes robot's \textit{actions}.
\end{itemize}
To our knowledge, our work is the first to experiment sensor fusion of raw camera and LiDAR data for proactive anomaly detection,
% / behavior prediction
and the above baselines are state-of-the-art methods for either anomaly detection tasks using unimodal perception signals or related tasks using multimodal perception signals. For a fair comparison, we implement all the backbone convolutional neural networks used across different methods for the camera image as the ResNet-18 pretrained on visual navigation task, as described in Section~\ref{subsec:training}. All methods are trained on the same dataset.

Quantitatively, we compare different methods using the following two metrics:
\begin{itemize}
\item
\textit{F1-score}: A comprehensive threshold-dependent index considering precision $P$ and recall $R$, which can be expressed as $2PR/(P+R)$. We set the threshold to be $0.5$, i.e., we declare a navigation failure if the predicted probability of failure is greater than that of being "normal" at a point in time.

\item
\textit{PR-AUC}:
%\footnote{PR-AUC focuses on the performance of the classifier on the minority class, thus is more suitable for imbalanced dataset than the area under the Receiver Operating Characteristic (ROC-AUC).}
A threshold-independent index indicating the area under the Precision-Recall Curve. PR-AUC describes the ability to distinguish between positive and negative samples for anomaly detection models.
\end{itemize}

We further employ the kernel density estimation~\cite{wkeglarczyk2018kernel} to fit probability density functions (pdfs) for normal and failure samples on the test set, respectively. We use a Gaussian kernel and apply the transformation trick~\cite{shalizi2013advanced} to make sure that the estimated pdfs have support on $[0,1]$.

\begin{figure}[t]
  \centering
  \begin{subfigure}[b]{0.65\linewidth}
    \includegraphics[width=\linewidth]{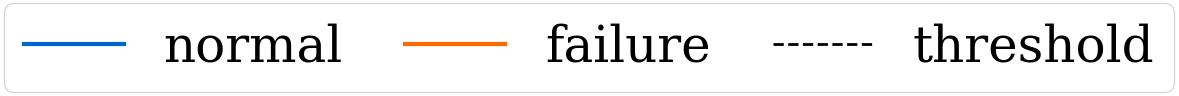}
  \end{subfigure}
  \par\smallskip
  \begin{subfigure}[b]{0.49\linewidth}
    \includegraphics[width=\linewidth]{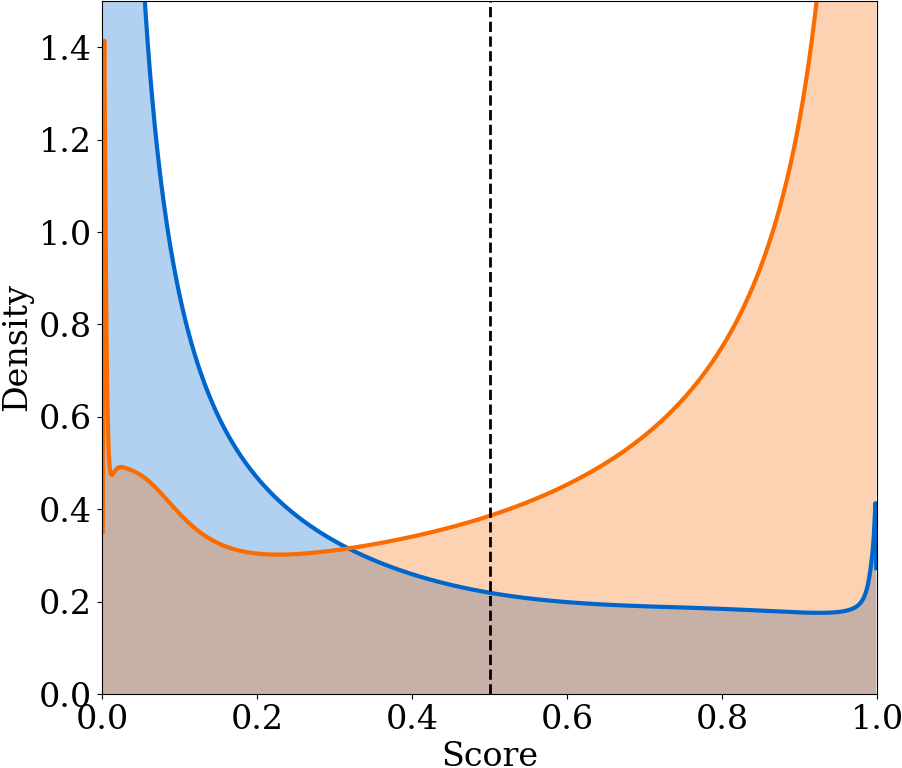}
    \caption{CNN-LSTM}
  \end{subfigure}
  \begin{subfigure}[b]{0.49\linewidth}
    \includegraphics[width=\linewidth]{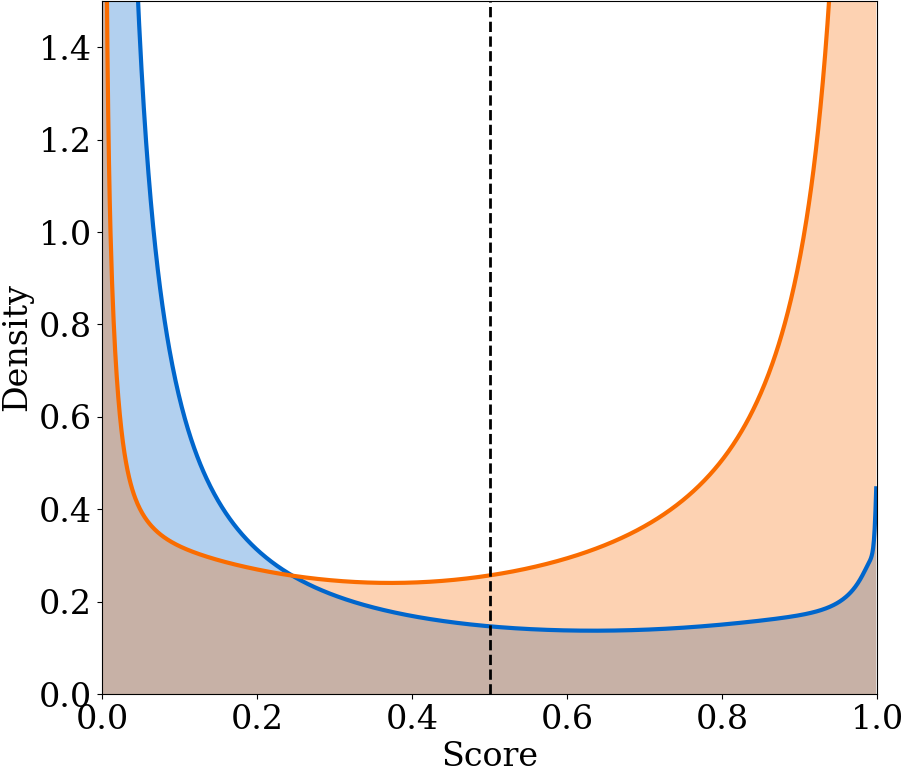}
    \caption{Cui et. al.}
  \end{subfigure}
  \par\smallskip
%   \begin{subfigure}[b]{0.47\linewidth}
%     \includegraphics[width=\linewidth]{Figures/dens_est_v3.png}
%     \caption{PAAD-camera}
%   \end{subfigure}
  \begin{subfigure}[b]{0.49\linewidth}
    \includegraphics[width=\linewidth]{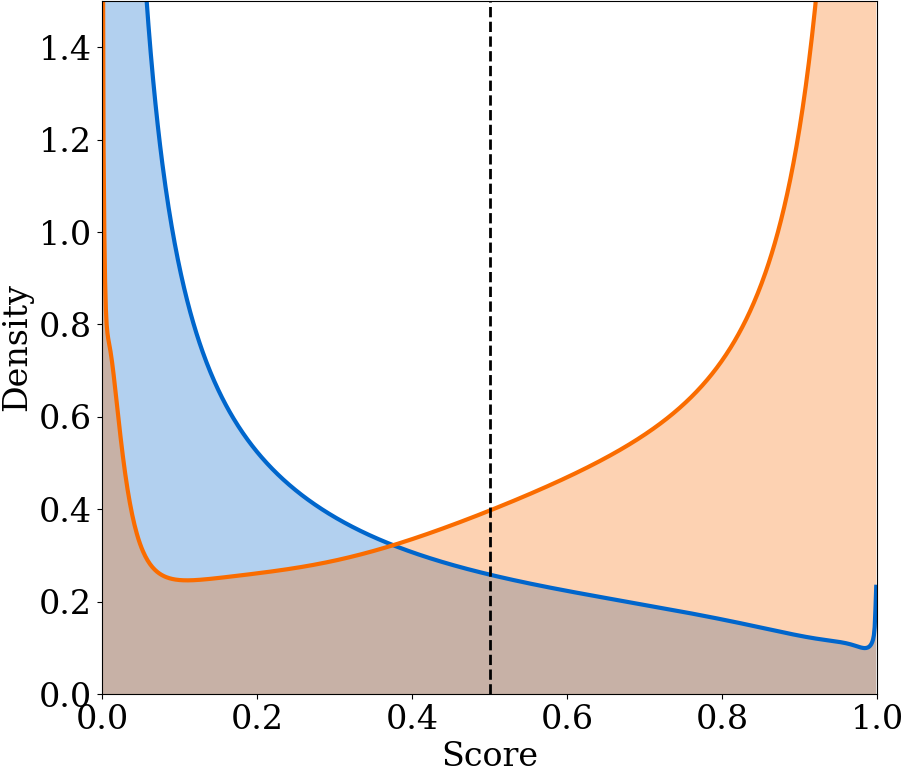}
    \caption{NMFNet}
  \end{subfigure}
  \begin{subfigure}[b]{0.49\linewidth}
    \includegraphics[width=\linewidth]{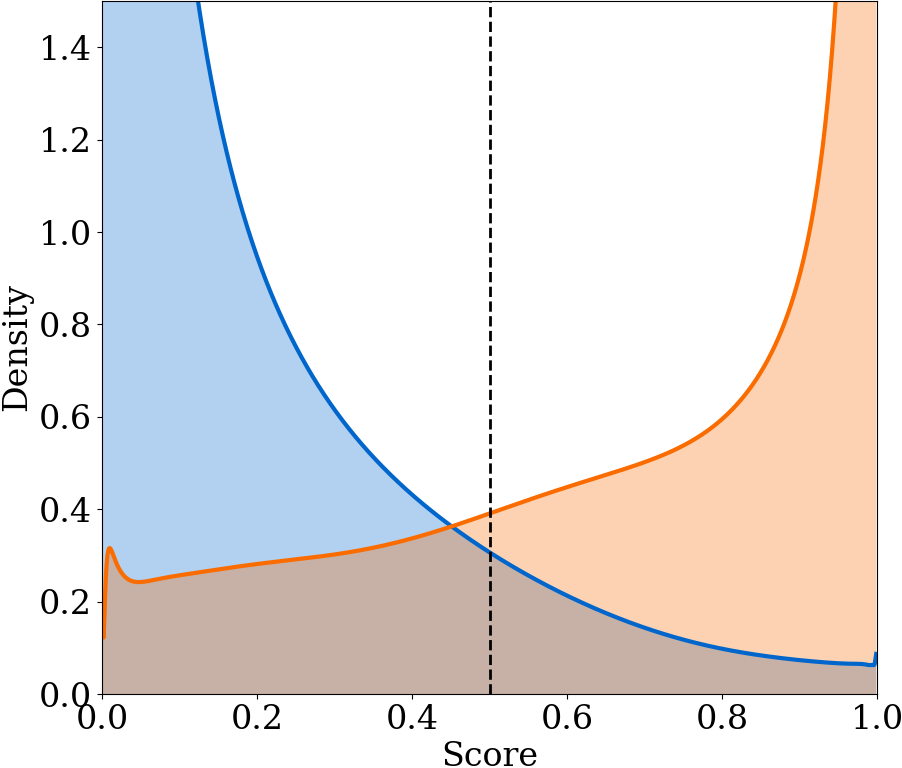}
    \caption{PAAD}
  \end{subfigure}
  \caption{\textbf{Distribution of predicted probability of failure for normal and failure points.} Upper parts of the pdfs are omitted. A clear separation of the two distributions is desired. An ideal anomaly detector should render the two pdfs as delta functions with impulses at $0$ and $1$ for normal and failure points, respectively.}
  \label{fig:density-estimation}
\end{figure}

The results are presented in Table~\ref{table:AD-performance} and Figure~\ref{fig:density-estimation}. As shown, PAAD achieves the best F1-score and highest PR-AUC with a large margin over other baselines. Although the CNN-LSTM model has been shown to have reliable anomaly detection performance for navigation tasks on sidewalks and off-road environments with large free space~\cite{kahn2021land,kahn2021badgr}, the method has not been shown to generalize well to harsh and cluttered field environments with limited open space. We argue that this is due to the fact that the control actions in such uncertain environments are high variance, making the network struggle with identifying true anomalous actions from noises. In fact, all the three baselines, which take the future control actions as input, make overconfident predictions for false positives and false negatives as shown in Figure~\ref{fig:density-estimation}. As a result, these three models in general show inferior F1-score and PR-AUC compared to PAAD, which makes use of the image representation of the planned path. Despite an additional sensor modality from LiDAR, NMFNet fails to provide a solid improvement over unimodal approaches, which highlights the importance of robust feature generator and fusion mechanism in highly uncertain environments.
\begin{table}[t]
  \begin{center}
    \caption{Anomaly detection performance with different methods}
    \label{table:AD-performance}
    \begin{tabular}{ l | c  c }
      \toprule
      Model & F1-score & PR-AUC \\
      \midrule
      \rule{-2.5pt}{2ex} CNN-LSTM~\cite{kahn2021land,kahn2021badgr} & $0.5352$ & $0.6988$ \\
      Cui et. al.~\cite{cui2019multimodal} & $0.5748$ & $0.7468$ \\
    %   PAAD-camera (ours) & $0.5700$ & $0.7784$ \\
      NMFNet~\cite{nguyen2020autonomous} & $0.5651$ & $0.7554$ \\
      \textbf{PAAD (ours)} & $\mathbf{0.6453}$ & $\mathbf{0.8281}$ \\
      \bottomrule
    \end{tabular}
  \end{center}
  \vspace{-3mm}
\end{table}

% By learning from the LiDAR signals, PAAD improves the anomaly detection performance over other baselines by a large margin. As shown in Figure~\ref{fig:density-estimation}, PAAD is more confident in detecting failures and tends to assign higher scores for anomalous points than PAAD-camera, thus leading to a higher PR-AUC. Additionally, with an extra sensor modality, PAAD is able to correctly identify normal cases where camera is occluded, which can otherwise be classified as anomalies by PAAD-camera. Such strengthened perception capability results in fewer false positives and thus a higher F1-score as shown in Figure~\ref{fig:density-estimation} and Table~\ref{table:AD-performance}, respectively.

\begin{figure*}[t]
  \centering
  \begin{subfigure}[c]{0.24\linewidth}
    \includegraphics[width=\linewidth]{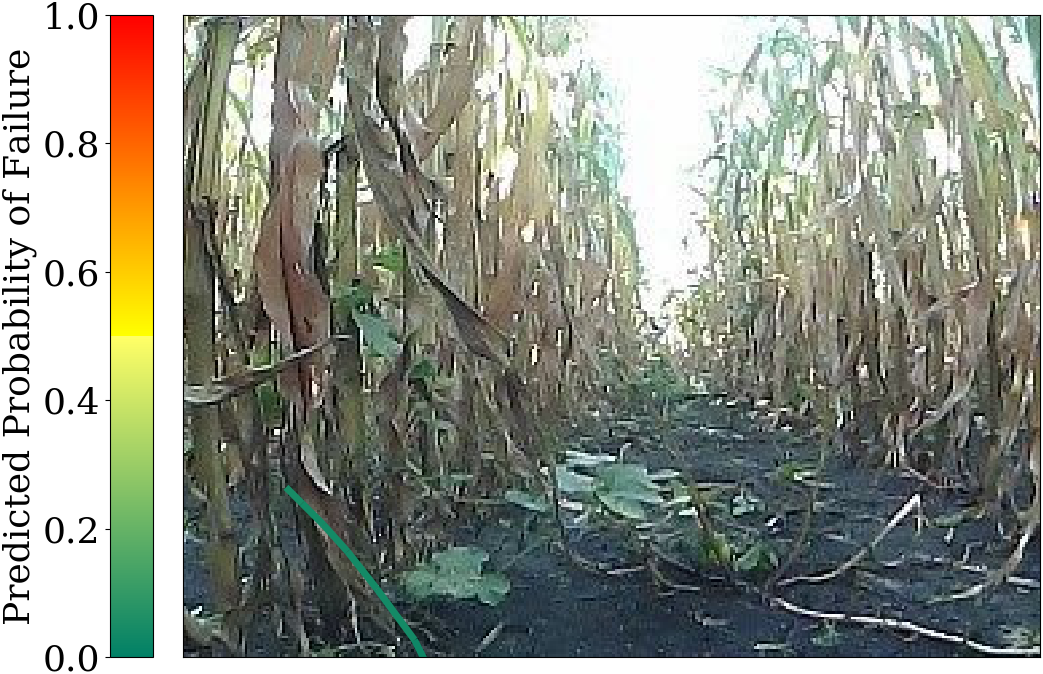}
  \end{subfigure}
  \begin{subfigure}[c]{0.2\linewidth}
    \includegraphics[width=\linewidth]{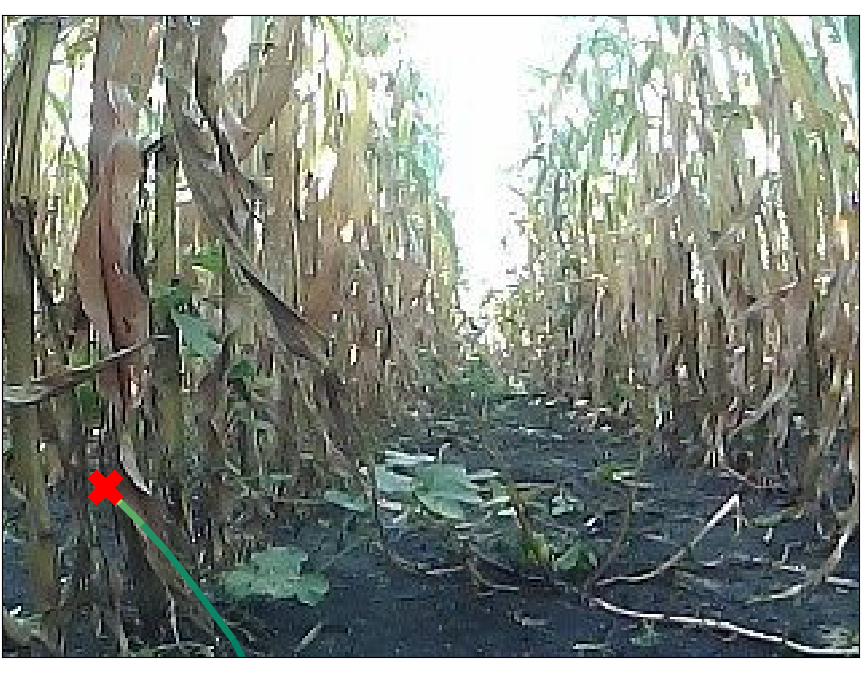}
  \end{subfigure}
%   \begin{subfigure}[c]{0.2\linewidth}
%     \includegraphics[width=\linewidth]{Figures/eg_1690_v3.png}
%   \end{subfigure}
  \begin{subfigure}[c]{0.2\linewidth}
    \includegraphics[width=\linewidth]{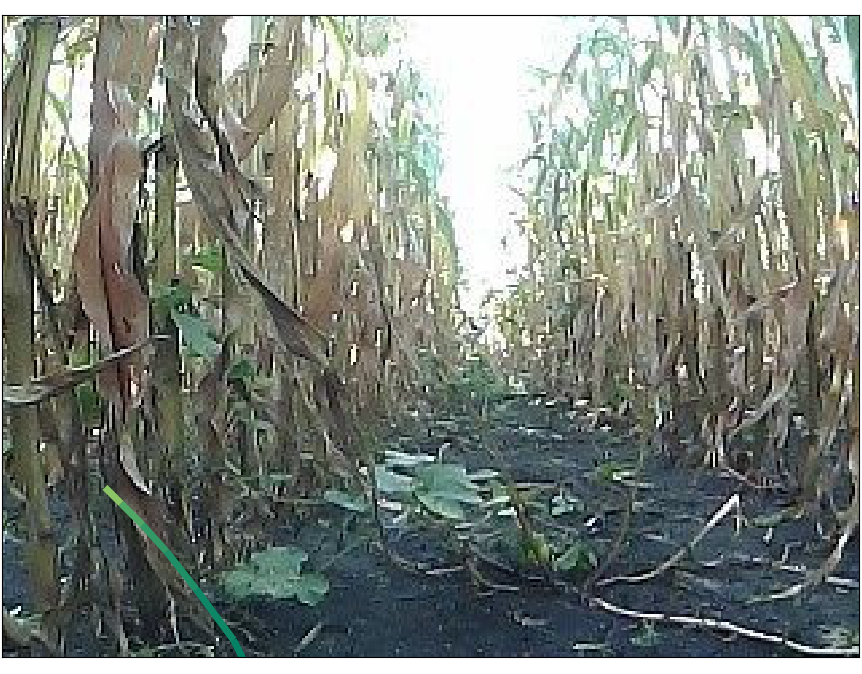}
  \end{subfigure}
  \begin{subfigure}[c]{0.2\linewidth}
    \includegraphics[width=\linewidth]{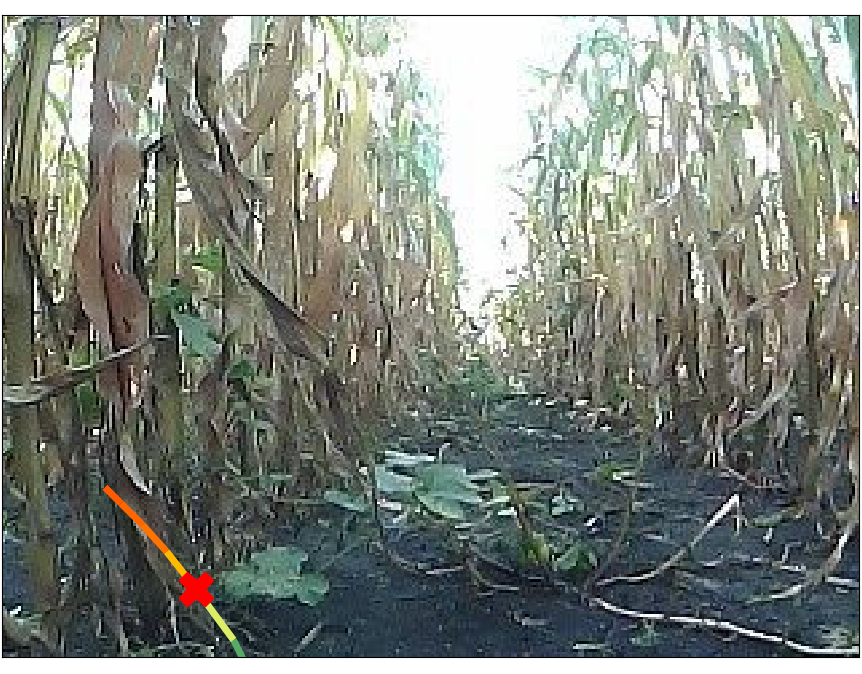}
  \end{subfigure}
  \begin{subfigure}[c]{0.11\linewidth}
    \includegraphics[width=\linewidth]{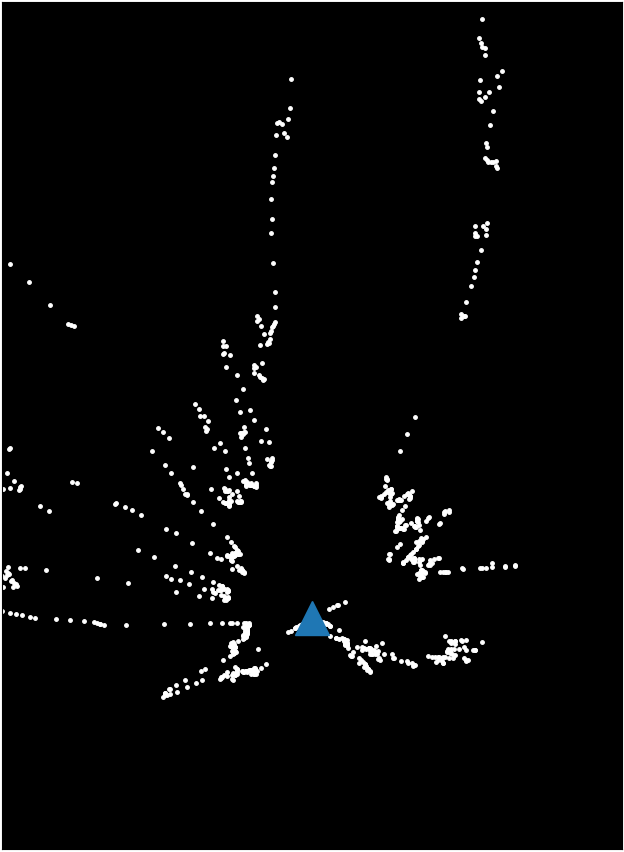}
  \end{subfigure}
  \begin{subfigure}[c]{0.24\linewidth}
    \includegraphics[width=\linewidth]{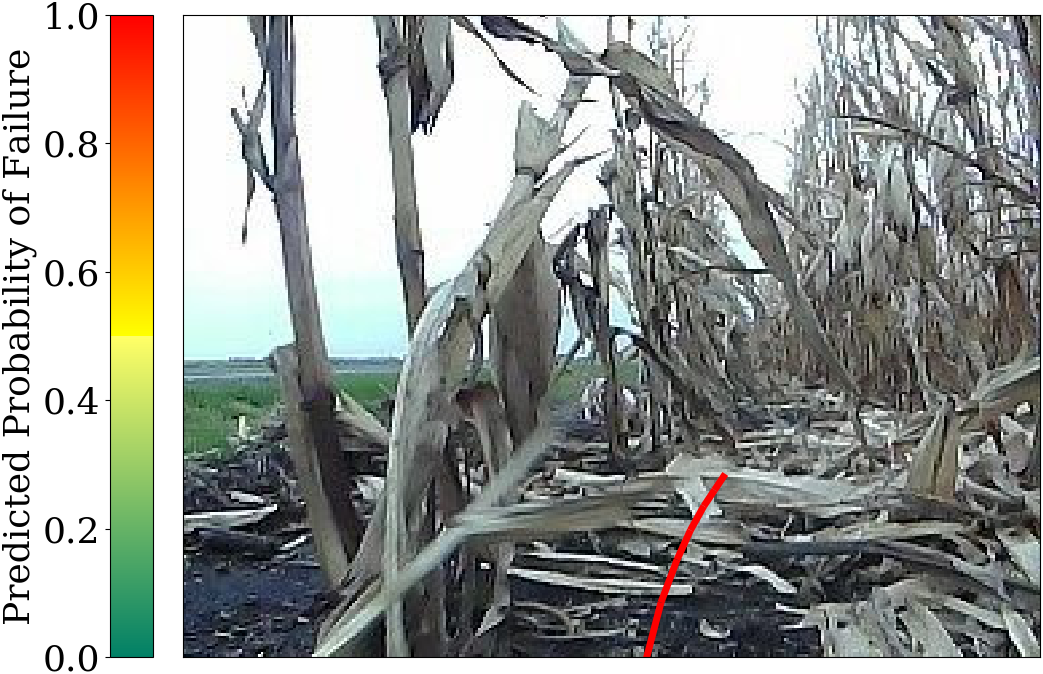}
  \end{subfigure}
  \begin{subfigure}[c]{0.2\linewidth}
    \includegraphics[width=\linewidth]{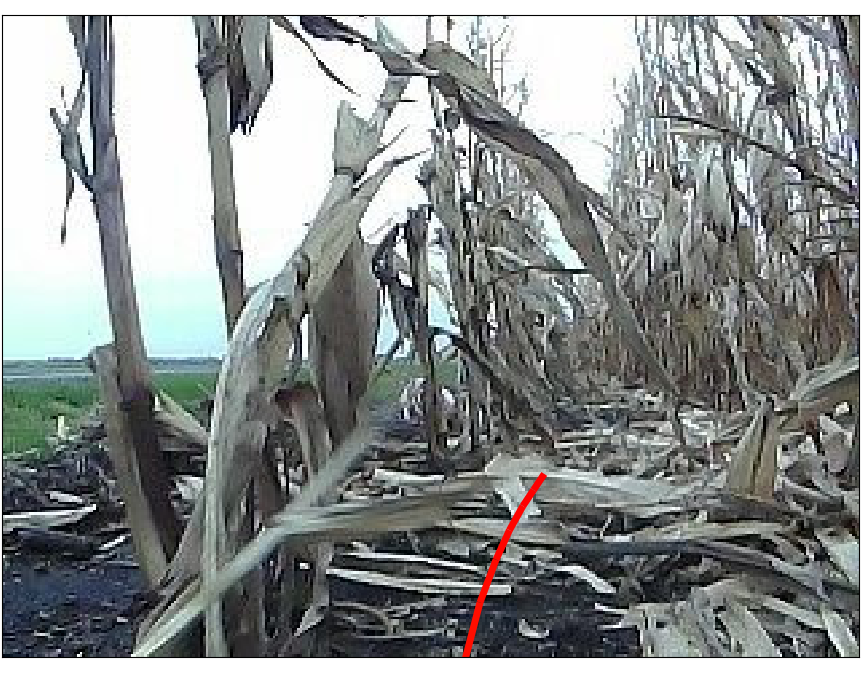}
  \end{subfigure}
%   \begin{subfigure}[c]{0.2\linewidth}
%     \includegraphics[width=\linewidth]{Figures/eg_5313_v3.png}
%   \end{subfigure}
  \begin{subfigure}[c]{0.2\linewidth}
    \includegraphics[width=\linewidth]{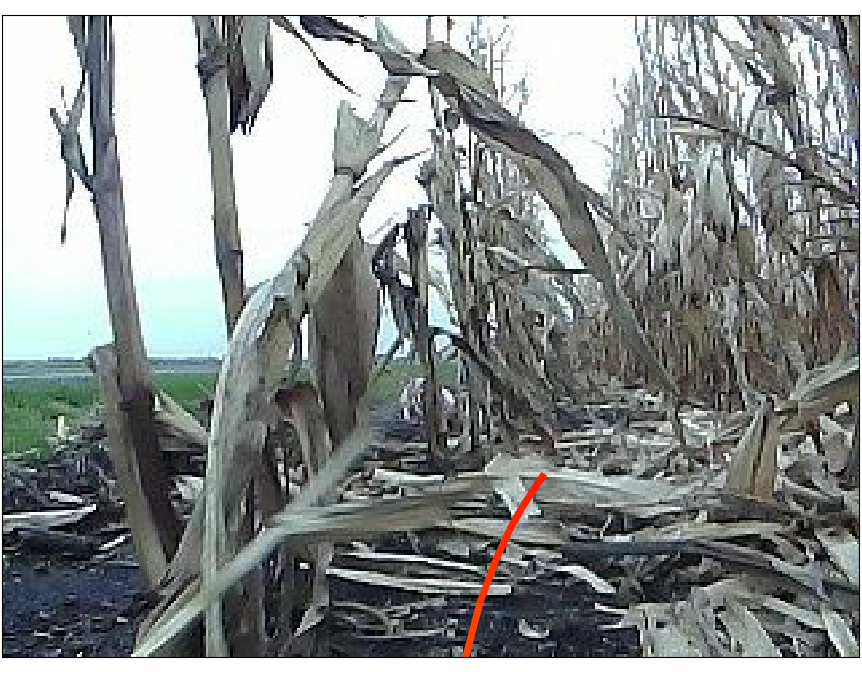}
  \end{subfigure}
  \begin{subfigure}[c]{0.2\linewidth}
    \includegraphics[width=\linewidth]{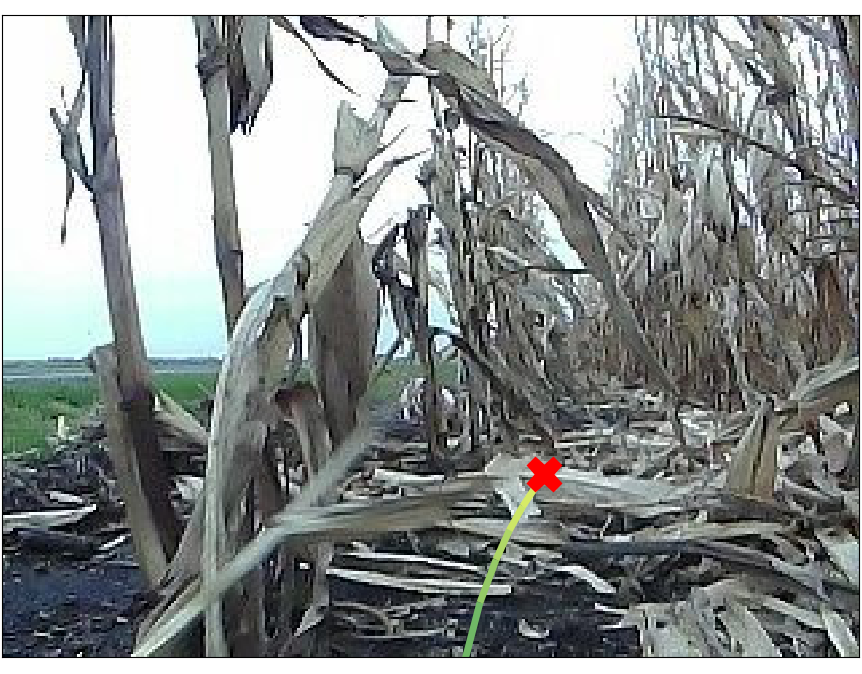}
  \end{subfigure}
  \begin{subfigure}[c]{0.11\linewidth}
    \includegraphics[width=\linewidth]{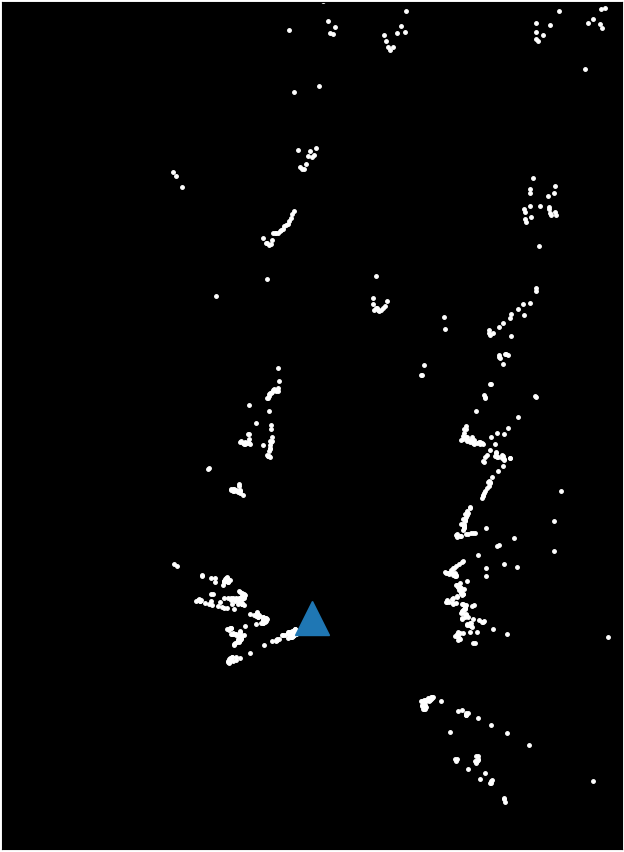}
  \end{subfigure}
  \begin{subfigure}[c]{0.24\linewidth}
    \includegraphics[width=\linewidth]{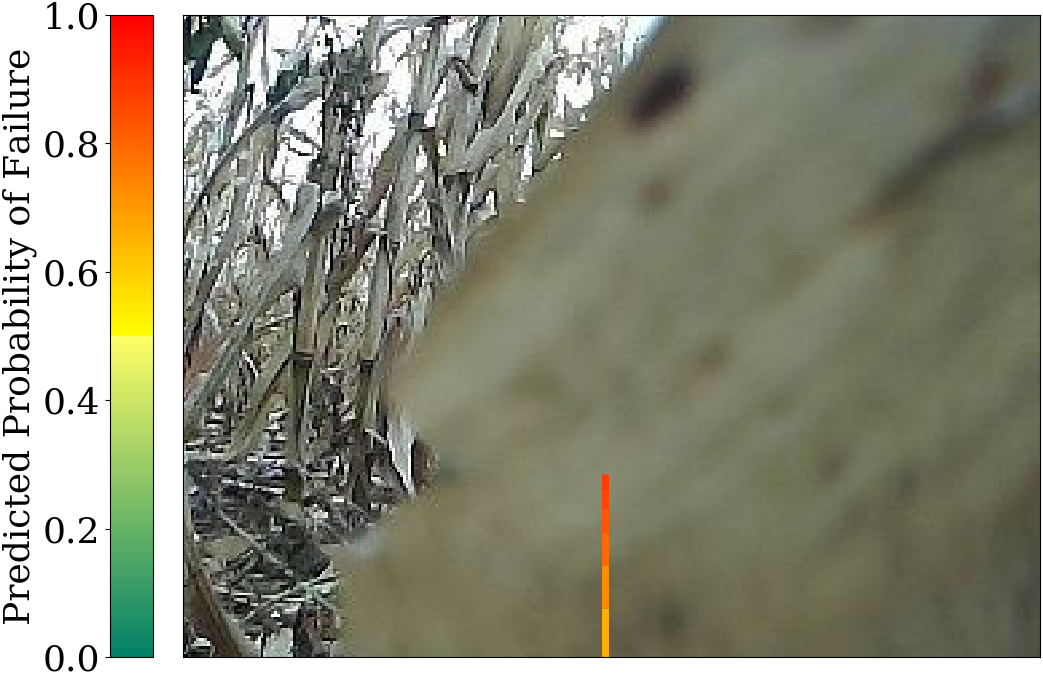}
    \caption{CNN-LSTM}
  \end{subfigure}
  \begin{subfigure}[c]{0.2\linewidth}
    \includegraphics[width=\linewidth]{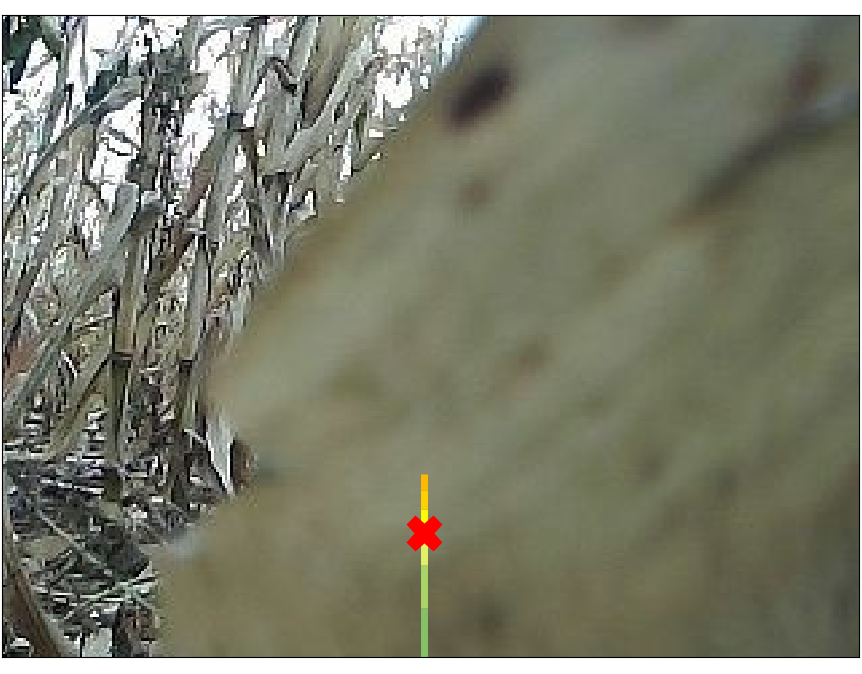}
    \caption{Cui et. al.}
  \end{subfigure}
%   \begin{subfigure}[c]{0.2\linewidth}
%     \includegraphics[width=\linewidth]{Figures/eg_7189_v3.png}
%     \caption{PAAD-camera}
%   \end{subfigure}
  \begin{subfigure}[c]{0.2\linewidth}
    \includegraphics[width=\linewidth]{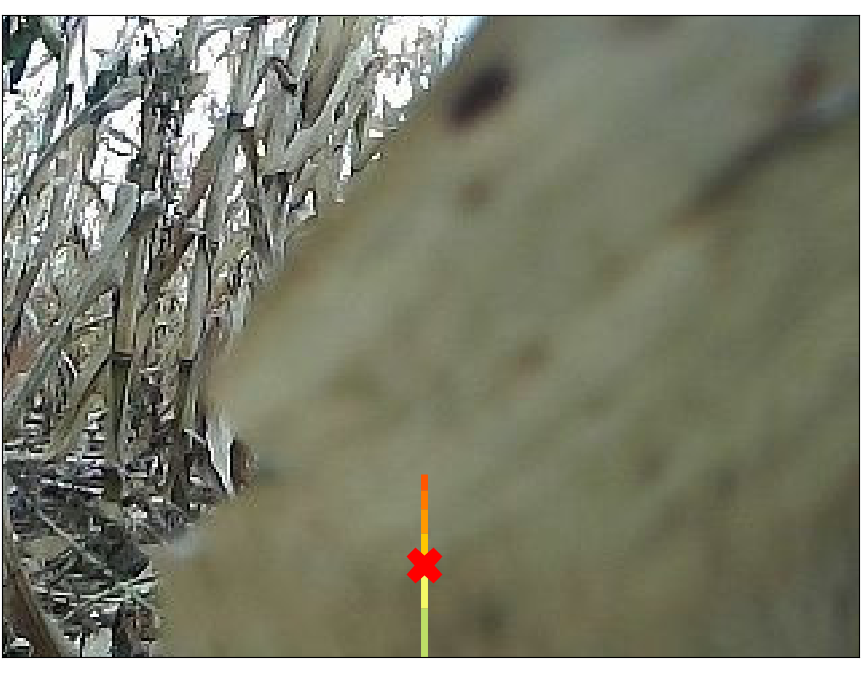}
    \caption{NMFNet}
  \end{subfigure}
  \begin{subfigure}[c]{0.2\linewidth}
    \includegraphics[width=\linewidth]{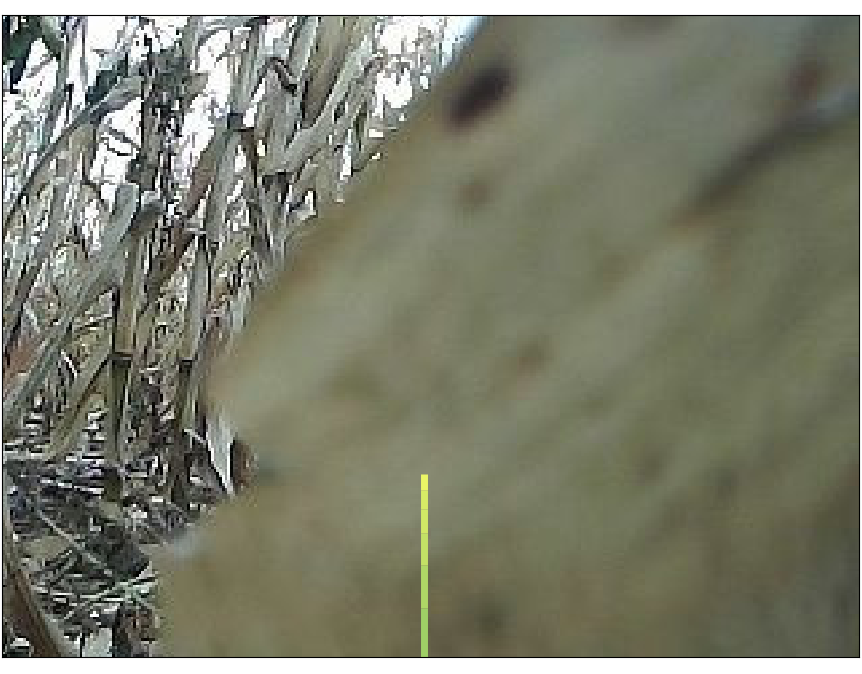}
    \caption{PAAD}
  \end{subfigure}
  \begin{subfigure}[c]{0.11\linewidth}
    \includegraphics[width=\linewidth]{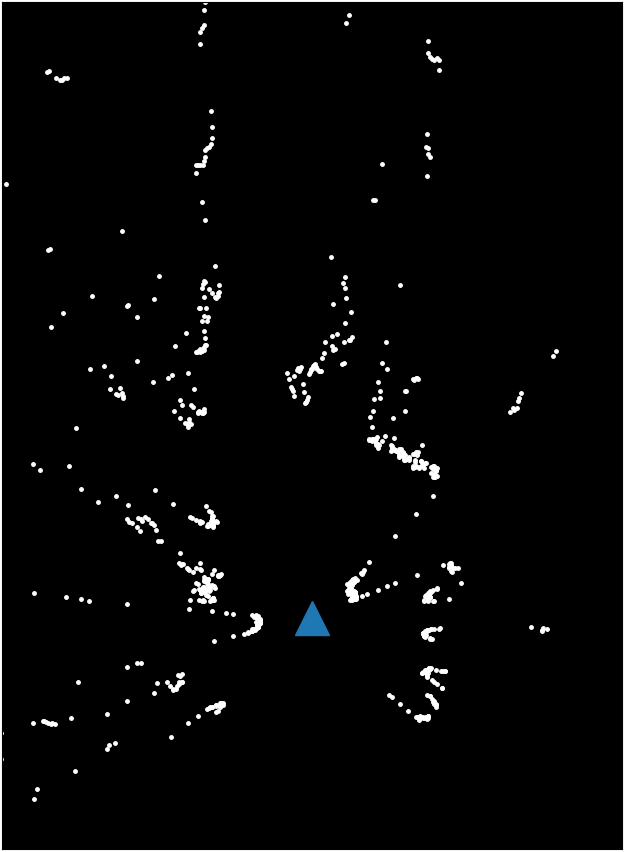}
    \caption{LiDAR data}
  \end{subfigure}
  \caption{\textbf{Qualitative comparisons of different methods in challenging environments.} CNN-LSTM, Cui et. al., and NMFNet are fed with planned control \textit{actions}, and the path shown above is only for visualization purpose. The LiDAR data is only used by NMFNet and PAAD for prediction. The red cross on the path denotes the first point at which the predicted probability of failure is over $0.5$.}
  \label{fig:examples}
  \vspace{-2mm}
\end{figure*}

\begin{figure}[t]
  \centering
  \begin{subfigure}[b]{0.52\linewidth}
    \includegraphics[width=\linewidth]{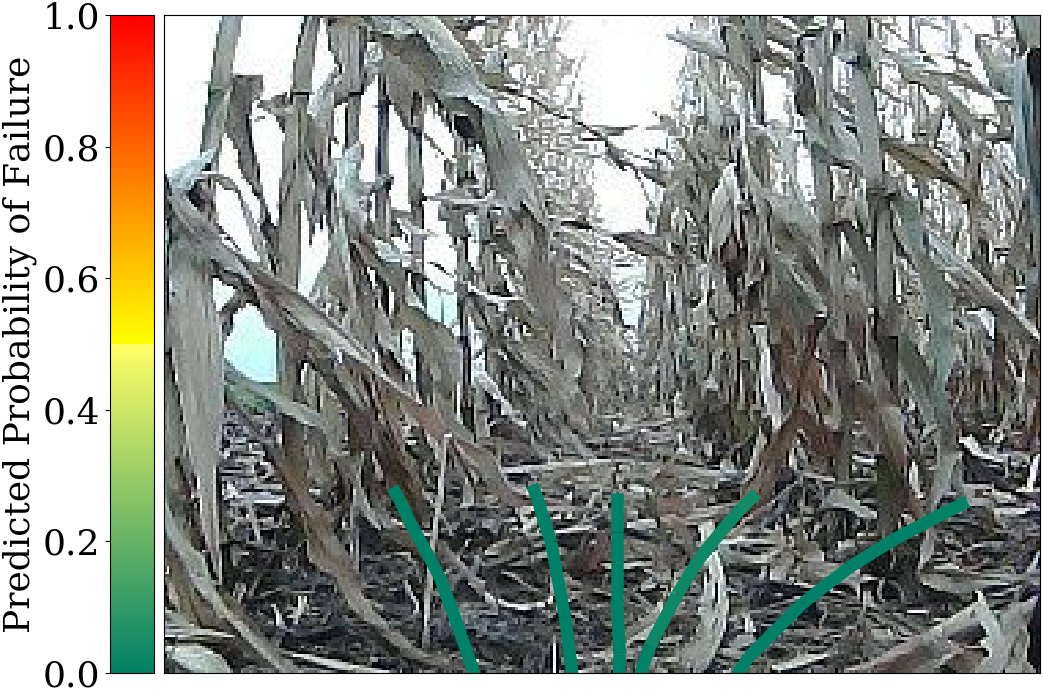}
    \caption{CNN-LSTM / Cui et. al. / NMFNet}
    \label{subfig:sample-path-baselines}
  \end{subfigure}
%   \begin{subfigure}[b]{0.52\linewidth}
%     \includegraphics[width=\linewidth]{Figures/sample_path_v1.png}
%     \caption{CNN-LSTM}
%     \label{subfig:sample-path-CNN-LSTM}
%   \end{subfigure}
%   \begin{subfigure}[b]{0.44\linewidth}
%     \includegraphics[width=\linewidth]{Figures/sample_path_v2.png}
%     \caption{Cui et. al.}
%     \label{subfig:sample-path-Cui}
%   \end{subfigure}
%   \begin{subfigure}[b]{0.52\linewidth}
%     \includegraphics[width=\linewidth]{Figures/sample_path_v3.png}
%     \caption{PAAD-camera}
%     \label{subfig:sample-path-PAAD-camera}
%   \end{subfigure}
%   \begin{subfigure}[b]{0.52\linewidth}
%     \includegraphics[width=\linewidth]{Figures/sample_path_vx.png}
%     \caption{NMFNet}
%     \label{subfig:sample-path-NMFNet}
%   \end{subfigure}
  \begin{subfigure}[b]{0.44\linewidth}
    \includegraphics[width=\linewidth]{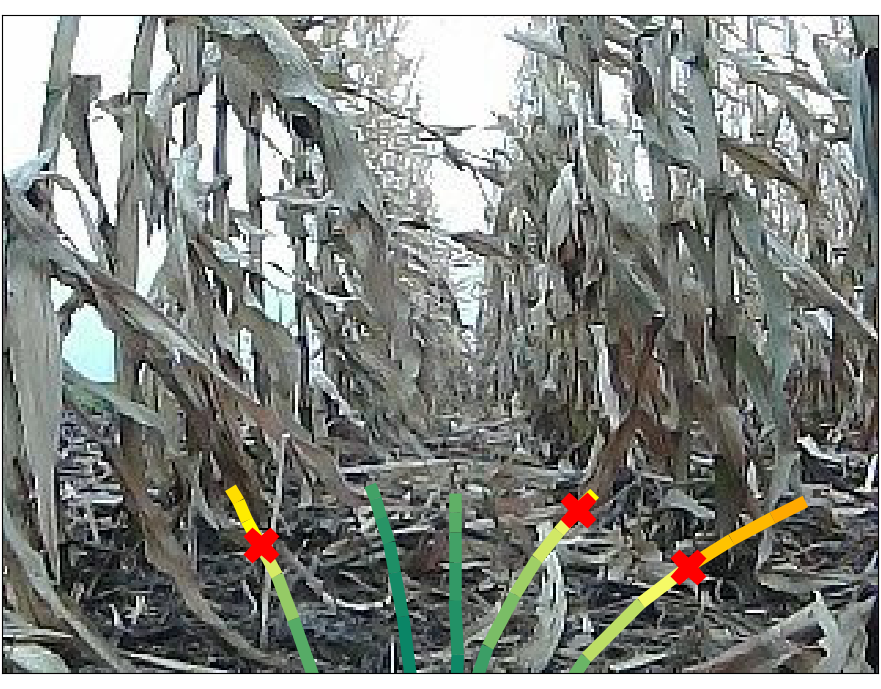}
    \caption{PAAD}
    \label{subfig:sample-path-PAAD}
  \end{subfigure}
  \caption{\textbf{Comparative study on AD performance with different sampled actions / paths.} All the three baselines generate similar probabilities of failure along the sampled paths and thus are condensed in one figure.}
  \label{fig:sample-path}
  \vspace{-2mm}
\end{figure}

Figure~\ref{fig:examples} shows the anomaly detection results of different methods in several challenging scenarios. In the first row, the LiDAR-based navigation algorithm falsely predict the orientation of the crop rows, making the robot take a left turn. As is further illustrated in Figure~\ref{fig:sample-path}, CNN-LSTM and NMFNet make the prediction of navigation failure merely based on the image without considering the future behavior, thus refusing to declare failures in such a clear image near the center line. Cui et. al.~\cite{cui2019multimodal} successfully detects a failure at the end of the path; however, the failure alert is too late to prevent the catastrophic collision. By contrast, the start time of the collision is more accurately predicted by PAAD. The second row shows a near-miss case where the robot manages to recover to the center line from the edge. Although PAAD falsely predicts a failure at the last point with a score of $0.52$, most part of the path is classified as normal correctly. However, all the other three methods generate overconfident scores for the entire path. The last row shows a normal case where the robot is tracking the center line while the camera is occluded by low-hanging leaves. The three baselines all failed while PAAD successfully distinguishes such normal behavior from an anomalous one.
%with additional information provided by LiDAR.

To further verify our hypothesis that noisy actions, as opposed to planned paths, hinder the network from learning useful features of robot's behavior, we feed an image and several sequences of actions / paths sampled from the test set through different models to predict probability of failure within the horizon. As shown in Figure~\ref{subfig:sample-path-baselines}, the three networks based on control actions always predict normal behaviors no matter how the future motion looks like, which indicates that the models are only making use of the image for anomaly detection. By contrast, PAAD can predict navigation failures based on the planned path, thus producing more promising results as shown in Figure~\ref{subfig:sample-path-PAAD}.

\begin{table}[t]
  \begin{center}
    \caption{Ablation study on PAAD}
    \label{table:ablation-study}
    \begin{tabular}{ l | c  c }
      \toprule
      Model & F1-score & PR-AUC \\
      \midrule
      \rule{-2.5pt}{2ex} LiDAR only & $0.5237$ & $0.6738$ \\
      camera only & $0.5700$ & $0.7784$ \\
      w/o MHA & $0.5941$ & $0.7902$ \\
      w/o reconstruction & $0.6218$ & $0.7997$ \\
      BEV & $0.6384$ & $0.7965$ \\
      \textbf{PAAD} & $\mathbf{0.6453}$ & $\mathbf{0.8281}$ \\
      \bottomrule
    \end{tabular}
  \end{center}
  \vspace{-3mm}
\end{table}

We further conduct an ablation study to reveal the benefit of different components in PAAD. The ablated versions of PAAD that we consider include:
\begin{inparaenum}[(1)]
\item
\textit{LiDAR only:} only the LiDAR pipeline is used to generate the observation features;
\item
\textit{camera only:} only the camera pipeline is used to generate the observation features;
\item
\textit{w/o MHA:} the residual MHA module is replaced with a simple MLP;
\item
\textit{w/o reconstruction:} the reconstruction branch in LiDAR pipeline is removed during training;
\item
\textit{BEV:} the planned path is projected to bird's eye view (BEV) instead of front view.
\end{inparaenum}
The results are summarized in TABLE~\ref{table:ablation-study}. With an extra sensor modality, PAAD is able to correctly identify normal cases where either camera or LiDAR is occluded, which can otherwise be classified as anomalies by LiDAR-only or camera-only. Such strengthened perception capability results in a higher F1-score and higher PR-AUC. The ablation study on other key components indicates the importance of each design choice to the overall performance of PAAD.

\subsection{Real-Time Test}
\label{subsec:real-time-test}
To test the ability of PAAD to alert the robot before executing an anomalous behavior, we further perform a real-time anomaly detection task on additional data\footnote{The test was performed on streaming data from rosbags to demonstrate the real-time capability of PAAD.}. In this experiment, the robot was driven by the vision-based navigation algorithm~\cite{sivakumar2021learned} on $1.3$ km of field trails, consisting of $750$m of common field environment and $550$m of densely weedy environment. Three and eight human interventions were required to reset the robot after an anomaly occurred in common and weedy environment, respectively.

We define the current \textit{anomaly score} as a linear combination of probabilities of failure within the prediction horizon:
\begin{equation}
s_t
=
\beta \sum_{k = 0}^{T - 1} \gamma^k y_{t + k},
\end{equation}
where $\gamma$ is a discount factor compensating the uncertainty in the future, and $\beta$ is a scaling factor ensuring that the summation $\sum_{k=0}^{T-1} \gamma^k$ equals $1$. At each time step $t$, we declare an anomaly if $s_t$ is greater than $0.5$.

To calibrate the difficulty of the task, we implement a LiDAR baseline for the real-time test. Given range measurements within the forward-facing $90^{\circ}$ field of view, we declare an anomaly if $85\%$ of the view is blocked by objects within $0.3$ meters. We also compare PAAD against a unimodal approach, Cui et. al.~\cite{cui2019multimodal}, and a multimodal approach, NMFNet~\cite{nguyen2020autonomous}, from Section~\ref{subsec:numerical-evaluation}. To increase the robustness against frequent occlusions of camera and LiDAR sensors in cluttered field environment, all the anomaly detectors declare an anomaly only when $3$ consecutive anomaly scores are over $0.5$. We implement all the methods at a frequency of $10$ Hz.

Table~\ref{table:real-time-test} summarizes the results. As shown, PAAD is able to detect anomalies reliably in both environments while maintaining a low false detection rate. On the contrary, the three baselines struggle with sensor occlusions and noisy actions in such cluttered and uncertain environments, thus frequently intervene the navigation system during the normal operation of the robot. Furthermore, we observe that PAAD is able to capture some rare failure modes, such as driving off the trail due to large gaps between crops. Scenarios in which PAAD failed usually contain dense weeds on the path and/or the robot executing near-miss maneuvers (see video). The detection of these anomalies could be potentially improved with additional data. Lastly, the reliable anomaly detection performance of the PAAD shown in the LiDAR-based navigation system (Section~\ref{subsec:numerical-evaluation}) and the vision-based navigation system (Section~\ref{subsec:real-time-test}) indicate that our method is agnostic to the underlying controller and can be applied to general systems that employ predictive control.
\begin{table}[t]
  \begin{center}
    \caption{Real-time test results}
    \label{table:real-time-test}
    \resizebox{\linewidth}{!}{%
    \begin{tabular}{ l | c  c  | c  c }
      \toprule
      \multirow{2}{*}{Method} & \multicolumn{2}{c|}{Common Field} & \multicolumn{2}{c}{Weedy Field} \\
                              & Anomaly Detected & False Detection & Anomaly Detected & False Detection \\
      \midrule
      LiDAR & $2/3$ & $10$ & $2/8$ & $>40$ \\
      Cui et. al. & $\mathbf{3/3}$ & $20$ & $\mathbf{8/8}$ & $>40$ \\
      NMFNet & $\mathbf{3/3}$ & $7$ & $6/8$ & $19$ \\
      \textbf{PAAD} & $\mathbf{3/3}$ & $\mathbf{1}$ & $\mathbf{7/8}$ & $\mathbf{8}$ \\
      \bottomrule
    \end{tabular}}
  \end{center}
  \vspace{-2mm}
\end{table}
\section{Conclusion}
In this work, we presented a proactive anomaly detection method for robot navigation in challenging field environment using multi-sensor signals. Our approach predicts the probability of future failure based on the planned path and the current sensor observation. By introducing a feature-level camera-lidar fusion, the detector successfully detected navigation failures in agricultural environment with higher F1-score and PR-AUC than other previous state-of-the-art methods. We also demonstrated the reliable anomaly detection performance of the PAAD with low false alarms in the real-time test. Although our method showed robustness in uncertain environments, false detection is unavoidable when both camera and LiDAR are blocked. Active perception, which encourages the robot to collect richer sensory signals through additional interaction with the environment, could decrease perception uncertainty in such cases of full sensor occlusion and would be a future work direction.
%Investigating how active perception can help correct false alarms would be a future work direction.
\section*{Acknowledgments}
This work was supported in part by the National Robotics Initiative 2.0 (NIFA\#2021-67021-33449) and AIFARMS through the Agriculture and Food Research Initiative (AFRI) grant no. 2020-67021-32799/project accession no.1024178 from the USDA/NIFA.
The robot platforms and data were provided by the Illinois Autonomous Farm and the Illinois Center for Digital Agriculture. 

%%%%%%%%%%%%%%%%%%%%%%%%%%%%%%%%%%%%%%%%%%%%%%%%%%%%%%%%%%%%%%%%%%%%%%%%%%%%%%%%

\bibliographystyle{IEEEtran}
\bibliography{BibFile}

\end{document}